\documentclass[10pt,twocolumn,letterpaper]{article}

\usepackage[pagenumbers]{cvpr} 

\usepackage[T1]{fontenc}
\usepackage{multirow}
\usepackage{tabularx}
\usepackage{float}
\usepackage{longtable}
\usepackage{pgfplots}
\pgfplotsset{compat=1.18}
\newsavebox{\resulttablebox}
\usepackage[skins,breakable]{tcolorbox}
\tcbset{analysisstatement/.style={
  enhanced,width=\linewidth,
  top=6pt,bottom=3pt,left=4pt,right=4pt,
  colback=blue!6!white,colframe=black,colbacktitle=black,
  fonttitle=\bfseries\small,
  attach boxed title to top left={yshift=-0.1in,xshift=0.1in},
  boxed title style={boxrule=0pt,colframe=white},
  before skip=7pt,after skip=5pt}}
\newtcolorbox{conjecturebox}[1]{analysisstatement,title={Conjecture #1}}
\newtcolorbox{findingbox}[1]{analysisstatement,title={Finding #1}}
\newtcolorbox{takeawaybox}[1]{
  enhanced,breakable,width=\linewidth,
  top=6pt,bottom=3pt,left=4pt,right=4pt,
  colback=blue!6!white,colframe=black,colbacktitle=black,
  fonttitle=\bfseries\small,title={Takeaway #1},
  attach boxed title to top left={yshift=-0.1in,xshift=0.1in},
  boxed title style={boxrule=0pt,colframe=white},
  before skip=7pt,after skip=5pt}

\usepackage{wrapfig}

\usepackage{algorithm}
\usepackage{algpseudocode}

\DeclareCaptionFont{dfccompact}{\fontsize{7.5}{8.5}\selectfont}
\DeclareCaptionFont{dfcmixseven}{\fontsize{7}{8}\selectfont}

\definecolor{cvprblue}{rgb}{0.21,0.49,0.74}
\usepackage[pagebackref,breaklinks,colorlinks,allcolors=cvprblue]{hyperref}
\usepackage{graphicx}
\usepackage{booktabs}
\usepackage{xcolor}
\usepackage{colortbl}
\definecolor{oursrow}{HTML}{EDA95F}
\definecolor{headerrow}{RGB}{200, 200, 200}
\definecolor{fullrow}{RGB}{238, 238, 238}
\def\confName{CVPR}
\def\confYear{2027}

\title{Omni-modal Compression can be Surprisingly Simple}

\author{
\textbf{Zhiyuan Liang}\textsuperscript{1}\quad
\textbf{Pengcheng Wang}\textsuperscript{2}\quad
\textbf{Jianyu Zhang}\textsuperscript{3}\\
\textbf{Jiashu Yang}\textsuperscript{3}\quad
\textbf{Wangbo Zhao}\textsuperscript{4}\quad
\textbf{Yang You}\textsuperscript{1}\\[4pt]
\textsuperscript{1}National University of Singapore\quad
\textsuperscript{2}USC\quad
\textsuperscript{3}Independent\quad
\textsuperscript{4}HKUST
}

\begin{document}
\maketitle
\suppressfloats[t]
\begin{abstract}
Omni-modal large language models (OLLMs) convert audio-visual clips into thousands of tokens, inducing substantial inference overhead.
Existing compression methods rely on heuristics such as key-frame selection, cross-modal token pairing, and additional scoring passes within the model. 
However, these procedures also introduce non-negligible bottlenecks that can hinder or even erase acceleration. 
We revisit if these techniques are necessary through a series of controlled experiments, finding that handcrafted metrics offer no consistent advantage over simple selection guided by the model's intrinsic attention saliency. 
These observations motivate Simple Omni Compression (SOC), a training-free framework that allocates token budgets uniformly across time, retains salient and diverse tokens within each frame, and uses attention from question tokens to refine the visual context at the middle layer.
We implement SOC with custom Triton kernels and integrate it with vLLM. 
Across audio-visual question answering, long-form generation, and multi-turn dialogue, SOC outperforms or matches six state-of-the-art compression methods across token retention ratios from 15\% to 45\%, with particularly strong performance under aggressive compression. 
SOC also translates reduced computation into practical acceleration, achieving a 1.66$\times$ inference speedup on LongShOT with 30B-A3B model at 35\% retention, while more elaborate methods can run slower than the uncompressed model.
Our code is available at \href{https://github.com/jerryliang24/Simple-Omni-Compression}{Simple-Omni-Compression}.
\end{abstract}    
\section{Introduction}
\label{sec:intro}
\begin{figure}[t]
    \centering
    \resizebox*{!}{6cm}{%
    \begin{minipage}{\linewidth}
    \centering
    \includegraphics[width=\linewidth]{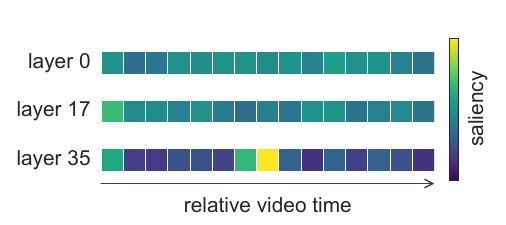}
    \par\vspace{5pt}
    \includegraphics[width=\linewidth,trim=0 5pt 0 0,clip]{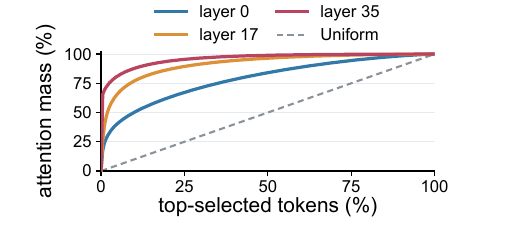}
    \end{minipage}%
    }
    \caption{
    \textit{Top} reveals OLLMs do not exhibit strong temporal selective bias until very deep layers.
    \textit{Bottom} indicates OLLM's token-level sparsity begins at a very early stage. 
    These suggest omni-modal compression can simply rely on model-derived saliency, without requiring elaborate, handcrafted inductive biases.}
    \label{fig:motivation}
\vspace{-2pt}
\end{figure}

\begin{figure*}[!t]
    \centering
    \includegraphics[width=\textwidth]{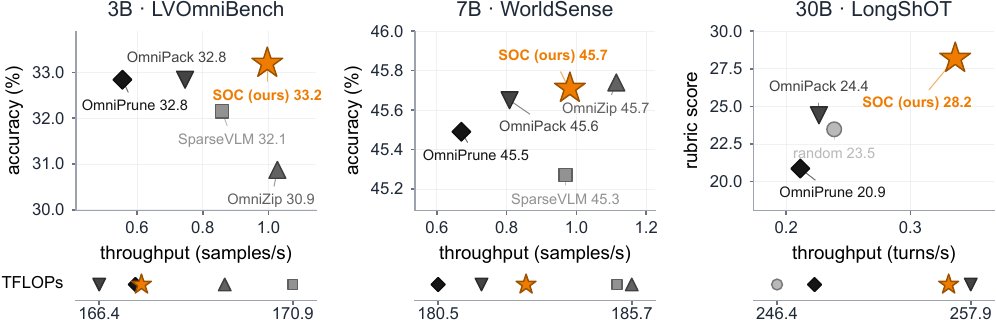}
    \caption{\textbf{Performance and efficiency illustration across three model sizes.} 
    At the 35\% reference retention tier, SOC achieves the highest task scores among the compared methods in all three settings while yielding top-tier throughput and theoretical FLOPs.}
    \label{fig:bubble}
\vspace{-1pt}
\end{figure*}

Advances in foundation models have sparked the emergence of Omni-modal Large Language Models (OLLMs)~\cite{wu2024nextgpt,li2024baichuan-o,shu2025audio,zhang2025llavavideo,xu2025qwen3omni,fu2026vita1_5}. 
Combining pretrained language backbone with modality encoders, these models integrate zero-shot question answering with visual and audio perception ability in one unified framework, supporting complex real-world understanding~\cite{zhou2025dailyomni, yang2025avut,fu2025videomme,tao2026lvomnibench, geng2025longvale, kurpath2025longshot}.

Despite their capacity, OLLMs are hindered by heavy computational burden, which stems from injecting massive modality tokens into LLM layers.
Prior works~\cite{chen2024fastv,zhang2025sparsevlm,tao2026omnizip,su2026omnipack,zhong2026omniprune,park2026omnidrop,yang2026omniselect} have explored omni-modal compression from multiple angles: attention-based saliency, cross-modal similarity, spatio-temporal redundancy, and external model guidance.
Though effective, these methods may introduce extra computations and lower the end-to-end speedup effect.
Moreover, modern hardware accelerations~\cite{kwon2023vllm,dao2024flashattention2} do not materialize attention score matrices, thus making end-to-end efficient and hardware-friendly algorithms crucial.

Evaluation gap remains another field to explore.
Existing methods primarily target multiple-choice QA~\cite{zhou2025dailyomni,yang2025avut,fu2025videomme,tao2026lvomnibench}, whose short outputs reveal little about decoding efficiency.
More challenging tasks, including long-form generation~\cite{geng2025longvale} and multi-turn conversation~\cite{kurpath2025longshot}, remain underexplored.
Prior evaluations also largely focus on 3B/7B models at 35\%--45\% token retention, leaving aggressive compression and scalability to larger models insufficiently studied.
Collectively, these limitations motivate broader evaluation across tasks, compression budgets, and model scales.

Despite the above, we find that \textit{omni context compression can be surprisingly simple}.
Leveraging the intrinsic representations learned by OLLM itself, we can avoid extra computation overhead and enable practical end-to-end inference acceleration.
Specifically, we uncover three important but often overlooked observations:

\textit{i) OLLMs exhibit little temporal preference in shallow layers.} 
It is often assumed that key events concentrate in a small subset of video frames, motivating the use of handcrafted metrics to locate the ``important'' frames~\cite{fu2025framefusion,shen2025longvu,tang2025aks}. 
However, Figure~\ref{fig:motivation} \textit{top} shows that attention across temporal windows is nearly identical in the shallow layers of OLLMs. 
For instance, the 8-th window is prominent at layer 35 but indistinguishable from others at layer 0.
Therefore, early heuristic frame selection may discard information needed in deeper layers, degrading performance.

\textit{ii) OLLMs exhibit token-level attention sparsity even in the earliest layers.}
Although temporal redundancy is hard to determine in shallow layers, OLLM's token-level sparsity is already pronounced.
As illustrated in Figure~\ref{fig:motivation} \textit{bottom}, a small fraction of tokens receives the majority of attention mass, even in the first LLM layer. 
This observation suggests that token-level redundancy can be identified early without other handcrafted priors and extra computations.

\textit{iii) Temporal attention becomes concentrated only after substantial model processing.}
Figure~\ref{fig:motivation} \textit{top} shows that attention shifts toward fewer temporal windows in deeper layers. 
This motivates revisiting token selection after contextual interactions, rather than performing massive token-pruning early. 
Together, these observations motivate our design: \textit{preserve temporal coverage during early compression, then refine the retained context within the LLM}.

Based on these observations, we propose \textbf{Simple Omni Compression (SOC)}, a training-free framework for reducing multi-modal context during OLLM inference.
SOC first retains salient and diverse tokens within each frame through model-intrinsic attention saliency and feature diversity, preserving temporal coverage with minimal extra overhead. 
After contextual interactions develop within the LLM, SOC further refines the visual context using middle-layer attention, enabling \textit{more aggressive compression}.
To translate token reduction into \textit{practical inference speedups}, we implement custom Triton kernels for saliency scoring, which accelerates by up to $9.8\times$ over naive PyTorch hooks and recomputation. 
For 30B-scale inference, we integrate SOC with vLLM~\cite{kwon2023vllm}, delivering a $1.66\times$ inference speedup at 35\% retention.
We further extend evaluation beyond short-answer QA to long-form generation and multi-turn conversation, demonstrating SOC's effectiveness across diverse tasks and compression budgets.

This paper's contribution is multi-fold:

\textbf{Simple yet scalable method}.
We introduce SOC, which combines encoder-based saliency and diversity selection with middle-layer refinement.
Despite its simplicity, SOC achieves a strong accuracy-latency trade-off across model scales and benchmarks as in Figure~\ref{fig:bubble}.

\textbf{Broader evaluation scenarios}.
We extend evaluation to long-form generation, multi-turn conversation, a 30B-A3B model, and more aggressive compression budgets, which are settings relatively underexplored in prior work.

\textbf{Hardware-compatible implementation}.
Our Triton kernels and vLLM integration complement SOC's algorithmic design, further boost practical inference acceleration.

\textbf{Comprehensive empirical study}.
Through a series of rigorous studies, we examine common compression assumptions and strategies to guide SOC's development.
Extensive experiments across tasks, model scales, and retention budgets assess SOC's quality and inference cost, while controlled ablations clarify each component's contribution and substantiate our design choices.

We also open-source our \href{https://github.com/jerryliang24/Simple-Omni-Compression}{code} to support future research.

\section{Analysis}
\label{sec:analysis}
We begin by examining the key factors in omni-modal context compression, providing empirical insights that guide the design of SOC.
Unless otherwise stated, the following results use Qwen2.5-Omni-3B on Daily-Omni benchmark.
In this section, a \emph{frame} denotes one visual temporal grid, comprising two consecutive video frames. 
All frame-level scores, budgets, and selection decisions operate on this unit.

\subsection{Visually-Salient Frame Hypothesis}
\label{sec:analysis-temporal}

\paragraph{Do OLLMs distinguish salient frames?} 
A widely-held belief is that a small subset of frames captures most of the salient information in video inputs. 
Motivated by this, many compression methods have devised heuristics to identify salient frames and allocate more computation accordingly. 
But does the OLLM itself preferentially attend to partial temporal windows when answering questions?

\noindent\textbf{\textit{Setup.}} We run Qwen2.5-Omni-3B on the five QA benchmarks in Section~\ref{sec:exp_setting} and extract the attention that visual tokens receive from the answer tokens $\mathcal{A}$ at every layer $\ell$.
For temporal window $\mathcal{W}_t$, we compute its saliency $S_t$ by averaging the attention scores across answer tokens and heads:
\begin{equation}
S_t^{(\ell)}=\frac{1}{|\mathcal{A}|}\sum_{i\in\mathcal{A}}\sum_{j\in\mathcal{W}_t}
\operatorname{softmax}_{j}\!\left(
\frac{q_{i}^{(\ell)}(k_j^{(\ell)})^\top}{\sqrt{d}}
\right).
\label{eq:temporal-window-saliency}
\end{equation}

\begin{figure}[t]
\centering
\includegraphics[width=\columnwidth,trim=0 7pt 0 0,clip]{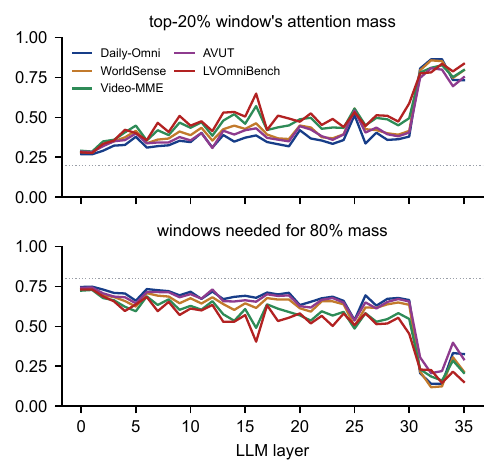}
\par\vspace{0pt}
\caption{Capturing majority of visual attention requires most temporal windows, while the top-ranked grids capture only a modest share, indicating broadly distributed attention in early layers.}
\label{fig:temporal-concentration}
\vspace{-2pt}
\end{figure}

Appendix~\ref{sec:appendix-attention-analysis} provides audio attention patterns, distributions across time and layers, and per-benchmark values.

\noindent\textbf{\textit{Results and Conclusions.}} Figure~\ref{fig:temporal-concentration} shows broadly distributed attention in early layers, with sharp concentration only from layer 31 onward.
Capturing 80\% of visual attention requires 60--75\% of temporal windows in layers 0--5, and this pattern persists through layers 6--30.
At the last layer, only 15--33\% of grids are needed, while the top 20\% capture 73--84\% of attention mass.
Thus, early-layer attention provides limited guidance for whole-frame selection.

\begin{takeawaybox}{1}
\textbf{Visual attention becomes sharply concentrated only in deep layers}, with broad distribution across temporal windows in shallow and intermediate layers.
\end{takeawaybox}
\begin{table}[H]
\centering
\footnotesize
\setlength{\tabcolsep}{2pt}
\renewcommand{\arraystretch}{1.5}
\begin{tabularx}{\columnwidth}{@{}p{0.23\columnwidth}p{0.30\columnwidth}>{\raggedright\arraybackslash}X@{}}
method & importance & note \\
\hline
encoder saliency & $\operatorname{TopK}_{j\in\mathcal{W}_t}(s_j)$ & TopK saliency for window $\mathcal{W}_t$ \\
saliency JS & $D_{\mathrm{JS}}(P_t,P_{t-1})$ & $P_t$: normalized saliency of $\mathcal{W}_t$ \\
feature change & $1-\cos(f_t,f_{t-1})$ & $f_t$: visual feature of the $\mathcal{W}_t$ \\
LLM saliency & $\sum_{j\in\mathcal{V}_t}\tilde{s}_j$ &  rank $\mathcal{W}_t$ via LLM saliency \\
\hline
\end{tabularx}
\caption{\textbf{Frame-scoring rules and intra-frame reference.}}
\label{tab:frame-importance-estimators}
\end{table}

\vspace{-20pt}
\begin{table}[H]
\centering\footnotesize
\setlength{\tabcolsep}{4pt}\renewcommand{\arraystretch}{1.05}
\newcommand{\FrameEntropyChange}{0.849}
\newcommand{\FrameEntropyKL}{0.965}
\newcommand{\FrameEntropyJS}{0.966}
\newcommand{\FrameEntropyDinoBAdjacent}{0.902}
\newcommand{\FrameEntropyDinoBWindow}{0.983}
\newcommand{\FrameEntropyDinoLAdjacent}{0.919}
\newcommand{\FrameEntropyDinoLWindow}{0.986}

\begin{tabularx}{\linewidth}[t]{Xrrr}
selection rule & $H_{\mathrm{norm}}$ & accuracy & $\Delta$ \\
\hline
\rowcolor{headerrow}base model & -- & 62.5 & -- \\
\rowcolor{fullrow}\multicolumn{4}{l}{\emph{token-level, encoder saliency}} \\
per-frame top-$k$ (uniform) & 1.000 & \textbf{61.7} & $-0.8$ \\
\rowcolor{fullrow}\multicolumn{4}{l}{\emph{whole-frame salient-frame selection}} \\
adjacent feature change (Omni encoder) & \FrameEntropyChange & 58.7 & $-3.8$ \\
saliency JS divergence & \FrameEntropyJS & 57.8 & $-4.7$ \\
saliency KL divergence & \FrameEntropyKL & 56.9 & $-5.6$ \\
DINOv3-B adjacent similarity & \FrameEntropyDinoBAdjacent & 58.6 & $-3.9$ \\
LLM layer-0 saliency & 0.970 & 55.6 & $-6.9$ \\
\hline
\end{tabularx}
\caption{\textbf{Uniform per-frame token allocation outperforms whole-frame selection.}
Per-frame top-$k$ saliency surpasses all whole-frame selection rules by 3.0--6.1 percentage points.}
\label{tab:frame-score-comparison}
\vspace{-2pt}
\end{table}

\paragraph{Does keeping visually-salient frames improve compression?}
Despite the limited guidance from OLLM attention, identifying salient temporal regions using other signals may be feasible.
For example, rapid changes in visual representations may indicate crucial video segments.
Such changes can be quantified using cosine distance between hidden representations or external visual features~\cite{shen2025longvu}.
We evaluate token fusion~\cite{fu2025framefusion, zhong2025aim} and token pairing~\cite{zhong2026omniprune, tao2026omnizip, su2026omnipack} in Appendices~\ref{sec:appendix-token-fusion} and~\ref{sec:analysis-coupling}, respectively.

\noindent\textbf{\textit{Setup.}} 
We evaluate encoder and LLM saliency, saliency-map JS/KL divergence, and cosine distance between Omni encoder/DINOv3-B features; Table~\ref{tab:frame-importance-estimators} summarizes the core estimators.
Before the OLLM backbone, we retain all visual tokens from the top-$k$ frames ranked by $S_t$ and discard the rest, 
keeping audio intact and adjusting $k$ to achieve 35\% total token retention.
We characterize frame-score distributions using normalized entropy $H_{\mathrm{norm}}$:
\begin{equation}
p_t = \frac{S_t}{\sum_{i=1}^{T} S_i},
\qquad H_{\mathrm{norm}} = -\frac{\sum_{t=1}^{T} p_t\log p_t}{\log T},
\label{eq:normalized-entropy}
\end{equation}

Higher entropy indicates a more uniform score distribution, for uniform distribution, $p_t = \frac{1}{T}$ and $H_{\mathrm{norm}}=1$. 
This assigns the same token budget for every temporal frame.

\noindent\textbf{\textit{Results.}} Table~\ref{tab:frame-score-comparison} reveals two patterns:
i) All frame-level selection rules incur substantial accuracy losses ($-3.8$ to $-6.9$).
Their high normalized entropy indicates limited differentiation between frames. 
Applying these flat scorers may remove information and degrade performance.
ii) Uniform per-frame top-$k$ selection achieves the highest accuracy, suggesting that token-level saliency preserves useful information without frame-level inductive biases.

\noindent\textbf{\textit{Conclusions.}} 
\textbf{Whole-frame selection offers no consistent advantage over intra-frame token selection.}
The evidence favors allocating tokens broadly throughout the sequence over focusing on certain frames.

\begin{table}[H]
\centering
\captionsetup{font=footnotesize,justification=raggedright,singlelinecheck=false}
\centering\footnotesize
\setlength{\tabcolsep}{1pt}\renewcommand{\arraystretch}{1.05}
\begin{tabularx}{\linewidth}{>{\hsize=.95\hsize\linewidth=\hsize\raggedleft\arraybackslash}X>{\hsize=.95\hsize\linewidth=\hsize\raggedleft\arraybackslash}X>{\hsize=1.2\hsize\linewidth=\hsize\raggedleft\arraybackslash}X>{\hsize=.95\hsize\linewidth=\hsize\raggedleft\arraybackslash}X>{\hsize=.95\hsize\linewidth=\hsize\raggedleft\arraybackslash}X}
$\rho$ & $\delta$ & $H_{\mathrm{norm}}$ & acc. & $\Delta$ \\
\hline
\rowcolor{headerrow}\multicolumn{2}{l}{base model} & -- & 62.5 & -- \\
\multicolumn{2}{l}{uniform} & 1.000 & 61.7 & $-0.8$ \\
\rowcolor{fullrow}\multicolumn{5}{l}{\textit{mild grading}} \\
0.4 & 0.1 & 0.996 & \textbf{61.9} & $-0.6$ \\
0.5 & 0.1 & 0.996 & 61.5 & $-1.0$ \\
0.3 & 0.2 & 0.987 & 61.2 & $-1.3$ \\
0.4 & 0.2 & 0.984 & 61.8 & $-0.7$ \\
0.5 & 0.2 & 0.983 & \textbf{61.9} & $-0.6$ \\
\rowcolor{fullrow}\multicolumn{5}{l}{\textit{strong grading}} \\
0.1 & 0.6 & 0.962 & 61.7 & $-0.8$ \\
0.2 & 0.6 & 0.924 & 61.5 & $-1.0$ \\
\rowcolor{fullrow}\multicolumn{2}{l}{whole-frame} & 0.657 & 58.7 & $-3.8$ \\
\hline\end{tabularx}
\captionsetup{font=footnotesize}
\caption{\textbf{Grading mitigates whole-frame pruning losses but offers no consistent gain over uniform allocation.}}
\label{tab:mild-temporal-allocation}

\end{table}

\paragraph{Can milder allocation help salient-frame selection?}

Preceding results show no consistent benefit from whole-frame decisions. 
Could a more balanced per-frame token budget while still favoring salient frames be more effective?

\noindent\textbf{\textit{Setup.}} At visual retention ratio $q$, we retain a fraction $d$ of tokens in top $\rho$ salient frames and $d-\delta$ tokens for the rest:
\begin{equation}
q = \rho d + (1-\rho)(d-\delta).
\label{eq:mild-allocation-budget}
\end{equation}
Thus, $d=q+(1-\rho)\delta$ and $d-\delta=q-\rho\delta$.
The allocation gap $\delta$ interpolates between uniform allocation ($\delta=0$) and whole-frame selection ($\delta=d$).

We use adjacent-frame cosine distance to select visual tokens within each frame, retaining 35\% of total tokens while preserving all audio tokens.
Allocation uniformity is measured by normalized entropy as in Equation~\ref{eq:normalized-entropy}.

\noindent\textbf{\textit{Results and Conclusions.}} \enlargethispage{\baselineskip}
Table~\ref{tab:mild-temporal-allocation} highlights that: 
i) Graded allocation limits accuracy losses within $-1.3$ to $-0.6$, versus $-3.8$ for whole-grid selection, supporting token coverage in non-salient frames.
ii) Neither grading strategy consistently improves on uniform selection: all settings fall within $-0.5$ to $+0.2$. 
These results favor preserving tokens across frames over prioritizing salient frames.

\begin{table}[H]
\centering
\captionsetup{font=footnotesize,justification=raggedright,singlelinecheck=false}
\centering\footnotesize
\setlength{\tabcolsep}{1pt}\renewcommand{\arraystretch}{1.05}
\begin{tabular*}{\linewidth}{@{\extracolsep{\fill}}lrrr@{}}
benchmark & full & uniform & graded \\
\hline
Daily-Omni & 62.5 & 61.7 & \textbf{61.9} \\
WorldSense & 46.4 & \textbf{45.5} & 45.2 \\
AVUT & 63.4 & \textbf{63.2} & 62.8 \\
Video-MME & 62.4 & \textbf{59.4} & 59.2 \\
LVOmniBench & 33.0 & 30.8 & \textbf{31.0} \\
\hline\end{tabular*}
\captionsetup{font=footnotesize}
\caption{\label{tab:keyframe-transfer}\textbf{Transfer experiments also favor uniform selection.} 
Across five QA benchmarks, graded allocation shows no consistent improvement.}

\end{table}

We further assess transfer across benchmarks using $\rho=0.4$, $\delta=0.1$ on WorldSense~\cite{hong2026worldsense}, AVUT~\cite{yang2025avut}, Video-MME~\cite{fu2025videomme}, and LVOmniBench~\cite{tao2026lvomnibench} and report in Table~\ref{tab:keyframe-transfer}.
The results also support \textbf{milder frame budget offers no consistent advantage over uniform selection.}

\begin{takeawaybox}{2}
\textbf{Salient-frame selection and enforced frame budget allocation offer no consistent advantage.}
Simple intra-frame token selection with uniform temporal coverage suffices for omni-modal compression.
\end{takeawaybox}

\subsection{Token Budget Allocation}
\begin{figure*}[!t]
    \centering
    \includegraphics[width=\textwidth]{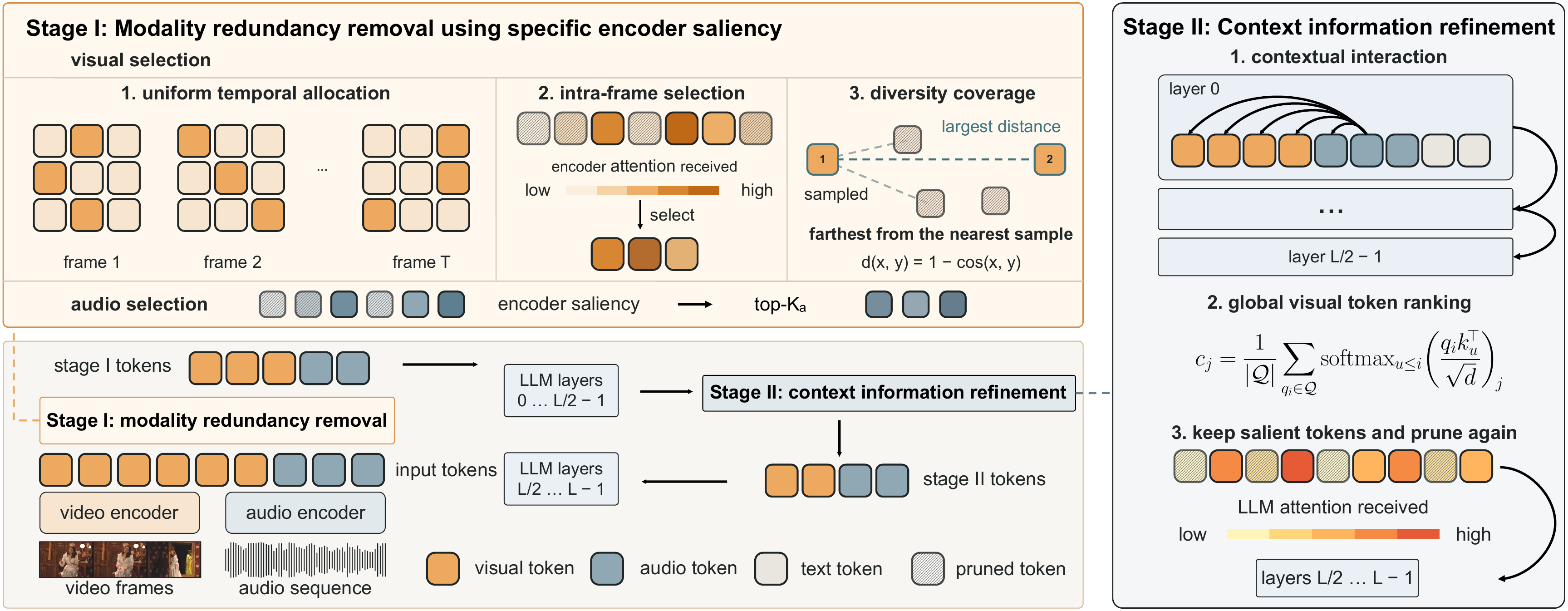}
    \caption{\textbf{Simple Omni Compression (SOC).}
    Stage I removes modality-specific redundancy using encoder saliency, uniform temporal allocation, and diversity coverage.
    Stage II refines the visual context via textual query attention after sufficient multimodal interaction.}
    \label{fig:method}
\end{figure*}

\label{sec:analysis-budget}

Beyond temporal distribution, we examine the audio token budget and how we balance token allocation across layers.

\noindent\textbf{Do we need to preserve most audio tokens?}
At 35\% retention, OmniPrune~\cite{zhong2026omniprune} retains 99.2\% of audio tokens and OmniZip~\cite{tao2026omnizip} retains 75\%.
Is preserving such a large fraction of audio tokens necessary?

\noindent\textbf{\textit{Setup.}} 
We sweep 25\%--55\% audio retention on Qwen2.5-Omni-3B/7B across five benchmarks while retaining all visual tokens.
\label{sec:analysis-audio-budget}
At 35\% total retention, we also reduce audio in OmniPrune and OmniZip and analyze performance.

\begin{table}[h]
\centering\scriptsize
\begin{subtable}{\columnwidth}
\centering
\setlength{\tabcolsep}{1.5pt}\renewcommand{\arraystretch}{1.15}
\begin{tabularx}{\columnwidth}{@{}lc*{4}{>{\centering\arraybackslash}X}@{}}
benchmark & base model & 25\% & 35\% & 45\% & 55\% \\
\hline
Daily-Omni & 62.5 & 61.3\,(-1.2) & 62.7\,(+0.2) & 62.9\,(+0.4) & 63.1\,(+0.6) \\
AVUT & 63.4 & 62.2\,(-1.2) & 63.4\,(0.0) & 63.9\,(+0.5) & 63.8\,(+0.4) \\
Video-MME & 62.4 & 62.3\,(-0.1) & 62.9\,(+0.5) & 62.8\,(+0.4) & 62.7\,(+0.3) \\
WorldSense & 46.4 & 44.8\,(-1.6) & 46.0\,(-0.4) & 46.3\,(-0.1) & 46.5\,(+0.1) \\
LVOmniBench & 33.0 & 32.5\,(-0.5) & 33.0\,(0.0) & 32.9\,(-0.1) & 32.6\,(-0.4) \\
\hline
\end{tabularx}
\captionsetup{font=footnotesize}
\caption{\textbf{Audio tokens can tolerate substantial pruning with moderate performance degradation}, with at most a $1.6$ accuracy drop.}
\label{tab:analysis-audio-fullvideo-3b}
\end{subtable}

\vspace{4pt}

\begin{subtable}{\columnwidth}
\centering
\setlength{\tabcolsep}{1.5pt}\renewcommand{\arraystretch}{1.15}
\begin{tabularx}{\columnwidth}{@{}lc*{4}{>{\centering\arraybackslash}X}@{}}
benchmark & base model & 25\% & 35\% & 45\% & 55\% \\
\hline
Daily-Omni & 63.9 & 61.7\,(-2.2) & 62.9\,(-1.0) & 63.1\,(-0.8) & 63.2\,(-0.7) \\
AVUT & 65.6 & 61.7\,(-3.9) & 64.1\,(-1.5) & 64.7\,(-0.9) & 65.2\,(-0.4) \\
Video-MME & 64.8 & 64.6\,(-0.2) & 64.9\,(+0.1) & 65.0\,(+0.2) & 65.0\,(+0.2) \\
WorldSense & 46.6 & 45.0\,(-1.6) & 45.8\,(-0.8) & 46.4\,(-0.2) & 46.4\,(-0.2) \\
LVOmniBench & 32.5 & 32.6\,(+0.1) & 32.6\,(+0.1) & 33.2\,(+0.7) & 32.4\,(-0.1) \\
\hline
\end{tabularx}
\captionsetup{font=footnotesize}
\caption{\textbf{Audio redundancy also holds on 7B model across benchmarks.}}
\label{tab:supp-fullvideo-audio}
\end{subtable}

\captionsetup{font=footnotesize}
\caption{\textbf{Audio pruning is robust across benchmarks and model scales.}}
\label{tab:audio-pruning-scales}
\vspace{-6pt}
\end{table}

\noindent\textbf{\textit{Results and Conclusions.}}
Tables~\ref{tab:analysis-audio-fullvideo-3b} and~\ref{tab:supp-fullvideo-audio} show that audio redundancy persists across benchmarks and model scales.
On 3B, retaining 25\% of audio tokens incurs at most a 1.6-point accuracy drop; the loss falls below 0.5 at 35\% and becomes negligible at 45\%.
Table~\ref{tab:baseline-audio-retention} further shows that reducing OmniPrune and OmniZip's audio token does not necessarily impair performance.
Thus, \textbf{substantial audio pruning is possible}, and preserving most audio tokens can overallocate the budget even in existing methods.

\begin{table}[H]\centering\footnotesize
\setlength{\tabcolsep}{3pt}\renewcommand{\arraystretch}{1.1}
\begin{tabularx}{\columnwidth}{Xccc}
 & video retention & audio retention & accuracy \\
\hline
\rowcolor{fullrow}OmniPrune  & 28.6\% & 99.2\% & 61.7 \\
audio 95\% & 29.0\% & 94.4\% & 61.8 \\
audio 75\% & 32.2\% & 75.0\% & 61.8 \\
audio 40\% & 35.1\% & 40.0\% & 61.3 \\
\hline
\rowcolor{fullrow}OmniZip  & 31.8\% & 75.0\% & 59.5 \\
audio 45\% & 34.9\% & 45.0\% & 59.6 \\
\hline
\end{tabularx}
\caption{\textbf{Existing methods may retain more audio tokens than necessary.}
Empirical results demonstrate reducing audio tokens does not necessarily degrade performance.}
\label{tab:baseline-audio-retention}
\end{table}

\vspace{-22pt}

\paragraph{Can contextualized representations inside OLLM improve token selection?}
Encoder saliency captures modality-intrinsic information, but may overlook task-relevant information.  
We examine whether OLLM attention saliency, which features multi-modal correlations and task dependencies, better guides token selection.
\begin{table}[H]
\captionsetup{font=footnotesize,justification=raggedright,singlelinecheck=false}
\centering\footnotesize
\setlength{\tabcolsep}{0.7pt}\renewcommand{\arraystretch}{1.05}
\begin{tabular*}{\linewidth}{@{\extracolsep{\fill}}ccccc@{}}
retention & $r_1\!\to\!r_2$ & stage I ($r_1$) & direct ($(r_1+r_2)/2$) & stage II ($r_2$) \\
\hline
15\% & $20\!\to\!10$ & 59.2 & 57.2 & 59.1 \\
20\% & $30\!\to\!10$ & 61.4 & 59.2 & 60.9 \\
35\% & $50\!\to\!20$ & 63.1 & 62.0 & 62.9 \\
45\% & $50\!\to\!40$ & 63.3 & 63.3 & 63.4 \\
\hline
\end{tabular*}
\caption{\textbf{Contextual refinement surpasses direct pruning}, highlighting the importance of task-relevant token selection. 
The listed token budgets are used as default setting throughout our experiments.}
\label{tab:analysis-stage-vs-direct}

\end{table}

\vspace{-16pt}

\noindent\textbf{\textit{Setup.}} 
We compare direct encoder-based pruning with a two-stage strategy: 
encoder-based selection (stage I) and attention-based refinement at the middle LLM layer (stage II), 
under matched layer-average retention.

\noindent\textbf{\textit{Results and Conclusions.}} Table~\ref{tab:analysis-stage-vs-direct} shows that two-stage refinement consistently outperforms direct pruning across 15\%--45\% retention, 
suggesting \textbf{contextualized attention improves task-relevant token selection beyond employing modality-intrinsic saliency alone}.

\begin{takeawaybox}{3}
\textbf{Effective compression requires balancing audio redundancy and contextual relevance.}
Audio tokens tolerate substantial pruning, while further contextual refinement exceeds pure modality saliency.
\end{takeawaybox}

\section{Method}
\label{sec:method}

Section~\ref{sec:analysis} motivates uniform temporal coverage and token selection via modality encoder's saliency.
Simple Omni Compression (SOC) (Figure~\ref{fig:method}) follows these rules in two training-free stages:
i) Modality-specific redundancy removal using encoder saliency.
ii) Further visual context refinement via intermediate-OLLM attention.

\subsection{Token Budgets}
\label{sec:method-budget}
Consider an audio-visual sequence consisting of $N_a$ audio tokens $\mathcal{U}$ and $N_v$ visual tokens $\mathcal{V}$, with $N=N_a+N_v$.
Given an overall retention ratio $r$ and an audio retention ratio $a$, we determine the total token budget $K$ and the corresponding audio and visual budgets, $K_a$ and $K_v$, as follows:
\begin{equation}
K\!=\!\operatorname{round}(rN),\;
K_a\!=\!\operatorname{round}(aN_a),\;
K_v\!=\!K-K_a.
\label{eq:soc-budget}
\end{equation}

For two-stage pruning, layers 0 to $L/2-1$ process $K_1=\operatorname{round}(r_{1}N)$ multi-modal tokens and layers $L/2$ to $L-1$ process $K_2=\operatorname{round}(r_{2}N)$.
The retention ratio we report is the average of $L$ layers: $\bar r=\frac{1}{L}\sum_{l=0}^{L-1} r^{(l)}=(r_{1}+r_{2})/2$.
In our experiment, we match all methods' layer-average token retention for fair comparison.

\subsection{Encoder Saliency}
\label{sec:method-saliency}
SOC ranks tokens by the attention they receive in the final layer of respective modality encoder.
For token $j$, we sum the attention over queries and average across heads:
\begin{equation}
e_j=\sum_{i\in\mathcal B(j)}
\operatorname{softmax}_{u\in\mathcal B(j)}
\left(\frac{q_i{k_u}^{\top}}{\sqrt{d}}\right)_j.
\label{eq:soc-encoder-score}
\end{equation}

Since Qwen's encoders use block-wise attention, we aggregate only over $\mathcal{B}(j)$: queries in the same block as $k_j$.
For visual encoder, each block corresponds to a temporal frame; regarding audio encoder, $\mathcal{B}(j)$ comprises audio tokens within a 2-second chunk.
Since FlashAttention-2 does not materialize attention matrices, we devise a GPU-accelerated Triton kernel that computes saliency scores tile by tile, reusing FlashAttention-2's row-wise log-sum-exp for normalization without producing the full attention matrix.
More details in Appendix~\ref{sec:appendix-triton}.

\subsection{Diversity Feature Coverage (DFC)}
\label{sec:method-stage1}
Motivated by Section~\ref{sec:analysis-temporal}, SOC preserves temporal coverage by spreading budget $K_v$ uniformly across $T$ grids. 
Each grid receives $k_t = \operatorname{round}(K_v/T)$ visual tokens.

Within each grid, SOC combines saliency with feature diversity. 
We first retain the $\operatorname{round}(\alpha k_t)$ tokens with the highest saliency $e_j$. 
DFC fills the remaining slots by iteratively selecting tokens with the largest cosine distance:
\begin{equation}
j^\star=\arg\max_{j\in\mathcal{R}_t\setminus\mathcal{D}_t}
\min_{u\in\mathcal{D}_t} \bigl[1-\cos(x_j,x_u)\bigr],
\label{eq:soc-dfc}
\end{equation}

where $\mathcal{R}_t$ is the pool left by saliency selection and $\mathcal{D}_t$ is the set selected so far by DFC. We initialize $\mathcal{D}_t$ with token $x_{j_0}$ that is farthest from its nearest neighbor in $\mathcal{R}_t$:
\begin{equation}
j_0=\arg\max_{j\in\mathcal{R}_t}
\min_{\substack{u\in\mathcal{R}_{t}, u\neq j}}
\bigl[1-\cos(x_j,x_u)\bigr].
\label{eq:dfc-init}
\end{equation}

In our default implementation, we select $\alpha=0.7$ of retained tokens using encoder saliency and the rest using DFC.
We initialize DFC with the token farthest from its nearest neighbor among the remaining tokens.
We ablate both $\alpha$ and the initialization strategy in Section~\ref{sec:ablation}.

\subsection{Progressive Pruning}

\paragraph{Modality redundancy removal.}
SOC selects tokens before the LLM using modality-encoder saliency.
Within each visual grid, it selects $\operatorname{round}(\alpha k_t)$ tokens by saliency and rest tokens with DFC.
Audio tokens are selected by global top-\(K_a\) ranking of the window-wise normalized scores in Equation~\ref{eq:soc-encoder-score}.
This stage exploits solely modality-intrinsic representations and is independent of the current question.

\paragraph{Contextual refinement of visual context.}
\label{sec:method-stage2}
After cross-modal interaction establishes semantic relationships, SOC further refines visual context via middle-layer attention, scoring visual token $j$ by the average attention from queries $\mathcal{Q}$ of the textual question (cross-head average omitted):
\begin{equation}
c_j=\frac{1}{|\mathcal{Q}|}\sum_{q_i\in\mathcal{Q}}\operatorname{softmax}_{u\leq i}\!\left(\frac{q_{i}{k_{u}}^{\!\top}}{\sqrt{d}}\right)_{\!j}.
\label{eq:soc-context-score}
\end{equation}

SOC retains $K_v^{(2)}=K_2-K_a$ visual tokens by $c_j$, while keeping all audio tokens from the first stage. 
Since audio occupies 78\% of input tokens on LongShOT~\cite{kurpath2025longshot}, we instead refine audio context at the middle layer. 
For multi-turn conversation, we share stage-I compression and apply stage-II independently for each turn of question.
We also ablate the refinement layer and the budget split between stages I and II in Appendices~\ref{sec:appendix-refinement-layer} and~\ref{sec:appendix-schedules}.

\section{Experiment}
\label{sec:exp}

\subsection{Settings}
\label{sec:exp_setting}
\begin{table*}[!t]
\centering\footnotesize

\setlength{\tabcolsep}{3pt}\renewcommand{\arraystretch}{0.96}
\begin{lrbox}{\resulttablebox}\begin{tabularx}{\textwidth}{l|*{2}{>{\centering\arraybackslash}X}|*{2}{>{\centering\arraybackslash}X}|*{2}{>{\centering\arraybackslash}X}|*{2}{>{\centering\arraybackslash}X}|*{2}{>{\centering\arraybackslash}X}}
method & \multicolumn{2}{c|}{Daily-Omni} & \multicolumn{2}{c|}{WorldSense} & \multicolumn{2}{c|}{Video-MME} & \multicolumn{2}{c|}{AVUT} & \multicolumn{2}{c}{LVOmniBench} \\
\rowcolor{headerrow}retained ratio & \multicolumn{1}{>{\columncolor{headerrow}[\tabcolsep][\dimexpr\tabcolsep+0.1pt\relax]}c}{35\%} & \multicolumn{1}{>{\columncolor{headerrow}[\dimexpr\tabcolsep+0.1pt\relax][\tabcolsep]}c|}{15\%} & \multicolumn{1}{>{\columncolor{headerrow}[\tabcolsep][\dimexpr\tabcolsep+0.1pt\relax]}c}{35\%} & \multicolumn{1}{>{\columncolor{headerrow}[\dimexpr\tabcolsep+0.1pt\relax][\tabcolsep]}c|}{15\%} & \multicolumn{1}{>{\columncolor{headerrow}[\tabcolsep][\dimexpr\tabcolsep+0.1pt\relax]}c}{35\%} & \multicolumn{1}{>{\columncolor{headerrow}[\dimexpr\tabcolsep+0.1pt\relax][\tabcolsep]}c|}{15\%} & \multicolumn{1}{>{\columncolor{headerrow}[\tabcolsep][\dimexpr\tabcolsep+0.1pt\relax]}c}{35\%} & \multicolumn{1}{>{\columncolor{headerrow}[\dimexpr\tabcolsep+0.1pt\relax][\tabcolsep]}c|}{15\%} & \multicolumn{1}{>{\columncolor{headerrow}[\tabcolsep][\dimexpr\tabcolsep+0.1pt\relax]}c}{35\%} & \multicolumn{1}{>{\columncolor{headerrow}[\dimexpr\tabcolsep+0.1pt\relax][\tabcolsep]}c}{15\%} \\
\hline
\rowcolor{fullrow}Qwen2.5-Omni-3B & \multicolumn{2}{c|}{62.5} & \multicolumn{2}{c|}{46.4} & \multicolumn{2}{c|}{62.4} & \multicolumn{2}{c|}{63.4} & \multicolumn{2}{c}{33.0} \\
random & 56.8 & 48.5 & 42.8 & 39.7 & 59.5 & 55.7 & 59.0 & 53.2 & 31.6 & 31.6 \\
DyCoke & 59.2 & 53.9 & 45.1 & N/A & 58.9 & N/A & 61.8 & 56.2 & 30.7 & N/A \\
FastV & 60.9 & 51.4 & 45.2 & 42.2 & 59.5 & 53.0 & 59.6 & 53.5 & 31.2 & 32.2 \\
SparseVLM & \underline{62.5} & 55.6 & 45.2 & 41.7 & 61.3 & 56.1 & 62.1 & 54.6 & 32.1 & \textbf{33.4} \\
OmniPrune & 61.7 & \underline{56.9} & 45.4 & 42.5 & \underline{61.8} & 57.0 & \underline{62.6} & \underline{59.1} & \underline{32.8} & \underline{33.2} \\
OmniPack & 59.6 & 52.1 & \underline{45.5} & 42.1 & 61.1 & \underline{58.4} & 61.5 & 57.0 & \underline{32.8} & 31.6 \\
OmniZip & 59.5 & 54.7 & \textbf{45.8} & \underline{43.7} & 58.7 & N/A & 61.4 & 58.0 & 30.9 & 31.7 \\
\rowcolor{oursrow}SOC (Ours) & \textbf{62.9} & \textbf{59.1} & \textbf{45.8} & \textbf{43.9} & \textbf{62.1} & \textbf{58.9} & \textbf{62.8} & \textbf{61.1} & \textbf{33.2} & \textbf{33.4} \\
\hline
\rowcolor{fullrow}Qwen2.5-Omni-7B & \multicolumn{2}{c|}{63.9} & \multicolumn{2}{c|}{46.6} & \multicolumn{2}{c|}{64.8} & \multicolumn{2}{c|}{65.6} & \multicolumn{2}{c}{32.5} \\
random & 56.7 & 48.0 & 43.2 & 39.0 & 62.7 & 58.5 & 60.4 & 54.5 & 31.0 & 30.5 \\
DyCoke & 59.9 & 54.7 & 45.4 & N/A & 60.3 & N/A & \textbf{64.9} & 60.2 & 29.4 & N/A \\
FastV & 61.8 & 54.5 & 44.2 & 39.5 & 61.1 & 52.0 & 63.0 & 53.4 & 31.7 & 30.3 \\
SparseVLM & 62.2 & 57.6 & 45.3 & \underline{43.2} & 63.1 & 56.9 & 64.5 & 57.9 & \textbf{32.1} & \textbf{31.3} \\
OmniPrune & \textbf{62.9} & \underline{58.9} & 45.5 & 42.2 & 63.5 & 60.6 & 64.6 & \underline{61.1} & 31.8 & \underline{31.0} \\
OmniPack & \underline{62.4} & 55.2 & \underline{45.6} & 41.0 & \underline{63.7} & \underline{61.5} & 63.2 & 59.2 & \underline{31.9} & 30.8 \\
OmniZip & 61.8 & 56.9 & \textbf{45.7} & 43.1 & 60.2 & N/A & 64.5 & 60.6 & 30.8 & 29.3 \\
\rowcolor{oursrow}SOC (Ours) & 62.2 & \textbf{59.1} & \textbf{45.7} & \textbf{43.7} & \textbf{64.9} & \textbf{62.0} & \underline{64.8} & \textbf{61.6} & \textbf{32.1} & 30.6 \\
\hline
\end{tabularx}\end{lrbox}
\ifdim\wd\resulttablebox>\textwidth\resizebox{\textwidth}{!}{\usebox{\resulttablebox}}\else\usebox{\resulttablebox}\fi
\caption{\textbf{Performance on omni-modal question answering benchmarks.} 
The \textbf{best} result is in bold, and the \underline{second-best} is underlined. 
N/A indicates that the target token budget is unreachable due to method-specific constraints. 
Our method achieves strong performance at standard 35\% retention ratio, and remains competitive under more aggressive compression with only 15\% tokens retained.}
\label{tab:performance}
\end{table*}

\paragraph{Benchmarks.}
We evaluate SOC on: omni-modal question answering (Daily-Omni~\cite{zhou2025dailyomni}, WorldSense~\cite{hong2026worldsense}, Video-MME~\cite{fu2025videomme}, AVUT~\cite{yang2025avut}, and LVOmniBench~\cite{tao2026lvomnibench}); 
long-form generation (LongVALE~\cite{geng2025longvale}); and multi-turn conversation (LongShOT~\cite{kurpath2025longshot}).
Details are provided in Appendix~\ref{sec:appendix-benchmarks}.

\paragraph{Baselines.}
We compare SOC with DyCoke~\cite{tao2025dycoke}, FastV~\cite{chen2024fastv}, SparseVLM~\cite{zhang2025sparsevlm}, OmniPrune~\cite{zhong2026omniprune}, OmniPack~\cite{su2026omnipack}, and OmniZip~\cite{tao2026omnizip}.
Since visual-only pruning prevents FastV and SparseVLM from reaching several target retention ratios, we extend their pruning mechanisms to audio tokens and jointly prune both modalities.
All experiments are conducted on NVIDIA H200 GPUs.

\paragraph{Implementation.}
Omni-QA benchmarks use BF16, FlashAttention-2, and greedy decoding with a maximum of eight generated tokens.
We sample videos at 2 FPS with at most 240 frames.
For Qwen3-Omni-30B-A3B~\cite{xu2025qwen3omni}, we use vLLM for inference.
LongVALE~\cite{geng2025longvale} uses greedy decoding with up to 1,024 new tokens and a repetition penalty of 1.1.
LongShOT~\cite{kurpath2025longshot} uses 32 uniformly sampled frames and generates up to 4,096 new tokens at a temperature of 0.1, with reference-answer history for each target turn.
We evaluate the random baseline with three seeds.
SOC retains 35\%/35\%/45\%/55\% of audio at 15\%/20\%/35\%/45\% overall retention and allocates the rest to video. 
Since audio takes 78\% of tokens on LongShOT, we retain 20\%/35\%/40\%/45\% of visual tokens, assign the rest to audio, and refine audio at the middle layer.

\subsection{General Omni-modal Question Answering}
\label{sec:exp_qa}

We evaluate SOC on five omni QA benchmarks using Qwen2.5-Omni-3B~\cite{Qwen2.5-Omni} and 7B and report accuracy at 35\% and 15\% token retention ratio, more results can be found in Appendix~\ref{sec:apx-complete-qa-results}. 
We can conclude from Table~\ref{tab:performance} that:

\textit{i) SOC is competitive across benchmarks and model sizes.}
At 35\% retention, SOC achieves the highest scores among compression methods across all benchmarks with the 3B model. 
For 7B scale, it leads on WorldSense, Video-MME, and LVOmniBench and ranks second on AVUT. 
These results show that SOC's simple design is competitive with more elaborate frameworks across tasks and scales.

\textit{ii) SOC is more robust under aggressive compression.}
At 15\% retention, SOC surpasses all baselines among most tasks at both model scales. 
These demonstrate that \textbf{simple design can remain robust under severe compute constraints}, supporting the building philosophy of SOC.

\begin{table}[!t]\centering\footnotesize
\setlength{\tabcolsep}{1pt}\renewcommand{\arraystretch}{1.04}
{\fontsize{7.5}{8.5}\selectfont\begin{tabularx}{\linewidth}{l>{\hsize=.8\hsize\linewidth=\hsize\centering\arraybackslash}X>{\hsize=1.2\hsize\linewidth=\hsize\centering\arraybackslash}X>{\hsize=.8\hsize\linewidth=\hsize\centering\arraybackslash}X>{\hsize=1.2\hsize\linewidth=\hsize\centering\arraybackslash}X}
method & FLOPs$\downarrow$ & latency (s)$\downarrow$ & accuracy $\uparrow$ & peak mem.$\downarrow$ \\
\hline\rowcolor{headerrow}base 3B model & 336.2\,T & 1.263 (1.00$\times$) & 33.0 & 12.18\,G \\
DyCoke & 167.3\,T & 0.859 (1.47$\times$) & 30.7 & 11.59\,G \\
FastV & 169.6\,T & 0.858 (1.47$\times$) & 31.2 & 11.59\,G \\
SparseVLM & 170.9\,T & 1.161 (1.09$\times$) & 32.1 & 11.59\,G \\
OmniPrune & 167.3\,T & 1.785 (0.71$\times$) & 32.8 & 11.59\,G \\
OmniPack & 166.4\,T & 1.320 (0.96$\times$) & 32.8 & 13.46\,G \\
OmniZip & 169.3\,T & 0.967 (1.31$\times$) & 30.9 & 11.59\,G \\
\rowcolor{oursrow}SOC (Ours) & 167.4\,T & 1.005 (1.26$\times$) & 33.2 & 11.59\,G \\
\hline\rowcolor{headerrow}base 7B model & 555.4\,T & 1.657 (1.00$\times$) & 32.5 & 22.12\,G \\
DyCoke & 237.0\,T & 0.978 (1.69$\times$) & 29.4 & 19.56\,G \\
FastV & 241.2\,T & 0.980 (1.69$\times$) & 31.7 & 20.72\,G \\
SparseVLM & 244.6\,T & 1.269 (1.31$\times$) & 32.1 & 20.72\,G \\
OmniPrune & 237.3\,T & 1.876 (0.88$\times$) & 31.8 & 19.56\,G \\
OmniPack & 238.5\,T & 1.582 (1.05$\times$) & 31.9 & 21.66\,G \\
OmniZip & 241.0\,T & 1.088 (1.52$\times$) & 30.8 & 19.94\,G \\
\rowcolor{oursrow}SOC (Ours) & 237.3\,T & 1.123 (1.48$\times$) & 32.1 & 19.56\,G \\
\hline\end{tabularx}}
\caption{\textbf{SOC yields promising efficiency gains while remaining capable}, whereas faster methods incur greater accuracy losses.}
\label{tab:efficiency-3B_7B}
\vspace{-10pt}
\end{table}

\begin{table*}[!t]\centering\footnotesize
\setlength{\tabcolsep}{1pt}\renewcommand{\arraystretch}{0.96}
\begin{subtable}[t]{.46\textwidth}\centering
\begin{tabularx}{\linewidth}{l>{\hsize=0.85\hsize\linewidth=\hsize\centering\arraybackslash}X>{\hsize=0.65\hsize\linewidth=\hsize\centering\arraybackslash}X>{\hsize=0.95\hsize\linewidth=\hsize\centering\arraybackslash}X>{\hsize=1.15\hsize\linewidth=\hsize\centering\arraybackslash}X>{\hsize=1.4\hsize\linewidth=\hsize\centering\arraybackslash}X}
method & SODA\_c & CIDEr & METEOR & \mbox{FLOPs (T)} & \shortstack{prefill (s)} \\
\rowcolor{fullrow}30B-A3B & 3.2 & 1.4 & 3.2 & 308.8 & 1.55 (1.00$\times$) \\
\hline\rowcolor{headerrow}\multicolumn{6}{l}{\textbf{45\%}} \\
random & 3.1 & \textbf{1.2} & \underline{3.0} & 137.0 & 0.90 (1.72$\times$) \\
OmniPrune & 3.1 & \textbf{1.2} & \underline{3.0} & 137.6 & 1.60 (0.97$\times$) \\
OmniPack & 2.9 & 0.3 & 2.3 & 139.6 & 1.49 (1.04$\times$) \\
OmniZip & \underline{3.2} & \underline{1.1} & \underline{3.0} & 138.5 & 0.97 (1.60$\times$) \\
\rowcolor{oursrow}SOC (Ours) & \textbf{3.3} & \underline{1.1} & \textbf{3.1} & 138.1 & 1.15 (1.34$\times$) \\
\hline\rowcolor{headerrow}\multicolumn{6}{l}{\textbf{35\%}} \\
random & \underline{3.0} & \underline{1.1} & \underline{2.9} & 115.3 & 0.80 (1.93$\times$) \\
OmniPrune & \underline{3.0} & \textbf{1.2} & \underline{2.9} & 114.7 & 1.55 (1.00$\times$) \\
OmniPack & 2.8 & 0.4 & 2.3 & 117.2 & 1.22 (1.27$\times$) \\
OmniZip & \underline{3.0} & 0.8 & \underline{2.9} & 117.4 & 0.88 (1.75$\times$) \\
\rowcolor{oursrow}SOC (Ours) & \textbf{3.1} & \textbf{1.2} & \textbf{3.0} & 119.2 & 1.10 (1.41$\times$) \\
\hline\rowcolor{headerrow}\multicolumn{6}{l}{\textbf{20\%}} \\
random & 2.8 & \underline{0.9} & \textbf{2.8} & 88.2 & 0.69 (2.24$\times$) \\
OmniPrune & \underline{2.9} & 0.7 & \underline{2.6} & 88.5 & 1.46 (1.06$\times$) \\
OmniPack & 2.7 & 0.4 & 2.2 & 88.5 & 0.91 (1.71$\times$) \\
OmniZip & \textbf{3.0} & 0.8 & \textbf{2.8} & 90.4 & 0.77 (2.02$\times$) \\
\rowcolor{oursrow}SOC (Ours) & \textbf{3.0} & \textbf{1.0} & \textbf{2.8} & 89.7 & 0.86 (1.79$\times$) \\
\hline\rowcolor{headerrow}\multicolumn{6}{l}{\textbf{15\%}} \\
random & 2.7 & \underline{0.8} & \textbf{2.8} & 80.7 & 0.64 (2.42$\times$) \\
OmniPrune & \underline{2.9} & 0.6 & 2.6 & 80.5 & 1.45 (1.07$\times$) \\
OmniPack & 2.7 & 0.3 & 2.2 & 80.8 & 0.83 (1.87$\times$) \\
OmniZip & \underline{2.9} & \textbf{1.0} & \textbf{2.8} & 82.3 & 0.72 (2.14$\times$) \\
\rowcolor{oursrow}SOC (Ours) & \textbf{3.0} & \textbf{1.0} & \underline{2.7} & 81.0 & 0.78 (1.99$\times$) \\
\hline\end{tabularx}
\caption{\textbf{LongVALE}: SOC adapts seamlessly to long-form generation, maintaining competitive quality across settings while saving prefill time.}\label{tab:30b-longvale}
\end{subtable}
\hfill
\begin{subtable}[t]{.52\textwidth}\centering
\begin{tabularx}{\linewidth}{l>{\hsize=0.75\hsize\linewidth=\hsize\centering\arraybackslash}X>{\hsize=0.75\hsize\linewidth=\hsize\centering\arraybackslash}X>{\hsize=0.85\hsize\linewidth=\hsize\centering\arraybackslash}X>{\hsize=0.55\hsize\linewidth=\hsize\centering\arraybackslash}X>{\hsize=0.7\hsize\linewidth=\hsize\centering\arraybackslash}X>{\hsize=1.4\hsize\linewidth=\hsize\centering\arraybackslash}X>{\hsize=2.0\hsize\linewidth=\hsize\centering\arraybackslash}X}
method & Rubric & Percep. & Reason. & Info. & Multi. & \mbox{FLOPs (T)} & \shortstack{latency (s)} \\
\rowcolor{fullrow}30B-A3B & 33.2 & 23.7 & 40.1 & 27.7 & 41.4 & 1009.9 & 4.95 (1.00$\times$) \\
\hline\rowcolor{headerrow}\multicolumn{8}{l}{\textbf{45\%}} \\
\rule[-1.20\dp\strutbox]{0pt}{\dimexpr1.20\ht\strutbox+1.20\dp\strutbox\relax}random & \underline{26.3} & 19.7 & \underline{31.3} & \underline{23.0} & \underline{31.3} & 324.7 & 3.78 (1.31$\times$) \\
\rule[-1.20\dp\strutbox]{0pt}{\dimexpr1.20\ht\strutbox+1.20\dp\strutbox\relax}OmniPrune & 20.6 & 17.1 & 25.4 & 18.8 & 20.9 & 323.7 & 4.78 (1.04$\times$) \\
\rule[-1.20\dp\strutbox]{0pt}{\dimexpr1.20\ht\strutbox+1.20\dp\strutbox\relax}OmniPack & 25.9 & \underline{19.9} & 29.8 & 20.8 & \textbf{33.4} & 338.6 & 5.04 (0.98$\times$) \\
\rowcolor{oursrow}\rule[-1.20\dp\strutbox]{0pt}{\dimexpr1.20\ht\strutbox+1.20\dp\strutbox\relax}SOC (Ours) & \textbf{28.9} & \textbf{23.8} & \textbf{36.5} & \textbf{25.8} & 29.7 & 327.1 & 3.21 (1.54$\times$) \\
\hline\rowcolor{headerrow}\multicolumn{8}{l}{\textbf{35\%}} \\
\rule[-1.20\dp\strutbox]{0pt}{\dimexpr1.20\ht\strutbox+1.20\dp\strutbox\relax}random & 23.5 & 18.1 & 27.3 & 19.6 & 28.9 & 246.4 & 4.19 (1.18$\times$) \\
\rule[-1.20\dp\strutbox]{0pt}{\dimexpr1.20\ht\strutbox+1.20\dp\strutbox\relax}OmniPrune & 20.9 & 16.7 & 23.8 & 16.4 & 26.6 & 248.6 & 4.73 (1.05$\times$) \\
\rule[-1.20\dp\strutbox]{0pt}{\dimexpr1.20\ht\strutbox+1.20\dp\strutbox\relax}OmniPack & \underline{24.4} & \underline{18.9} & \underline{27.6} & \underline{20.7} & \underline{30.5} & 257.9 & 4.42 (1.12$\times$) \\
\rowcolor{oursrow}\rule[-1.20\dp\strutbox]{0pt}{\dimexpr1.20\ht\strutbox+1.20\dp\strutbox\relax}SOC (Ours) & \textbf{28.2} & \textbf{23.2} & \textbf{33.8} & \textbf{24.6} & \textbf{31.3} & 256.6 & 2.98 (1.66$\times$) \\
\hline\rowcolor{headerrow}\multicolumn{8}{l}{\textbf{20\%}} \\
\rule[-1.20\dp\strutbox]{0pt}{\dimexpr1.20\ht\strutbox+1.20\dp\strutbox\relax}random & 20.5 & 15.9 & 23.2 & 16.7 & 26.1 & 155.1 & 3.67 (1.35$\times$) \\
\rule[-1.20\dp\strutbox]{0pt}{\dimexpr1.20\ht\strutbox+1.20\dp\strutbox\relax}OmniPrune & 21.1 & 15.7 & 25.1 & 15.5 & 28.0 & 154.7 & 4.69 (1.06$\times$) \\
\rule[-1.20\dp\strutbox]{0pt}{\dimexpr1.20\ht\strutbox+1.20\dp\strutbox\relax}OmniPack & \underline{23.1} & \underline{17.0} & \underline{25.5} & \underline{18.4} & \textbf{31.4} & 157.7 & 5.03 (0.98$\times$) \\
\rowcolor{oursrow}\rule[-1.20\dp\strutbox]{0pt}{\dimexpr1.20\ht\strutbox+1.20\dp\strutbox\relax}SOC (Ours) & \textbf{23.7} & \textbf{18.2} & \textbf{25.9} & \textbf{19.4} & \underline{31.3} & 158.2 & 2.72 (1.82$\times$) \\
\hline\rowcolor{headerrow}\multicolumn{8}{l}{\textbf{15\%}} \\
\rule[-1.20\dp\strutbox]{0pt}{\dimexpr1.20\ht\strutbox+1.20\dp\strutbox\relax}random & 20.9 & 15.8 & 23.8 & 16.6 & 27.6 & 131.6 & 3.01 (1.64$\times$) \\
\rule[-1.20\dp\strutbox]{0pt}{\dimexpr1.20\ht\strutbox+1.20\dp\strutbox\relax}OmniPrune & 20.3 & \underline{16.3} & \textbf{25.6} & 16.1 & 23.2 & 130.9 & 4.66 (1.06$\times$) \\
\rule[-1.20\dp\strutbox]{0pt}{\dimexpr1.20\ht\strutbox+1.20\dp\strutbox\relax}OmniPack & \underline{21.6} & 16.2 & 23.6 & \underline{16.9} & \underline{29.7} & 133.8 & 3.60 (1.38$\times$) \\
\rowcolor{oursrow}\rule[-1.20\dp\strutbox]{0pt}{\dimexpr1.20\ht\strutbox+1.20\dp\strutbox\relax}SOC (Ours) & \textbf{22.6} & \textbf{17.7} & \underline{24.7} & \textbf{17.8} & \textbf{30.2} & 132.8 & 2.58 (1.92$\times$) \\
\hline\end{tabularx}
\caption{\textbf{LongShOT}: Our method scales well to multi-turn conversation, with highest rubric score at every budget and significant latency reduction.}\label{tab:30b-longshot}
\end{subtable}
\caption{\textbf{Scaling SOC to 30B.} SOC maintains competitive generation quality on LongVALE and LongShOT, extending its capacity to long-form generation and complex multi-turn interaction while offering desirable efficiency.
OmniZip is omitted on LongShOT because minimum achievable audio-visual retention fails to reach the target budgets under the audio-dominant setting.}
\label{tab:30b-long}
\vspace{-10pt}
\end{table*}

\subsection{Efficiency Analysis}
\label{sec:exp_efficiency}

To compare SOC's efficiency gains with other baselines, we report inference latency, FLOPs, accuracy, and peak memory usage at 35\% retention using LVOmniBench for both 3B and 7B.
Results on all benchmarks and token budgets are provided in Appendix~\ref{sec:appendix-efficiency}.
Table~\ref{tab:efficiency-3B_7B} shows:

\textit{i) SOC translates token compression into practical inference gains.}
On LVOmniBench, SOC reduces FLOPs by 50\%/57\% and achieves $1.26\times$/$1.48\times$ speedups on 3B/7B. 
Despite similar FLOPs, OmniPrune and OmniPack take 1.785 s and 1.320 s, exceeding base 3B model's 1.263 s.
This highlights \textbf{simple compression method with minimal extra overhead is key to practical efficiency gains}.

\textit{ii) SOC offers a strong quality-latency balance.}
\enlargethispage{\baselineskip}
SOC achieves the highest accuracy at both scales, exceeding base 3B model by $0.2$ and trailing 7B model by $0.4$. 
It runs faster than half or more baselines, while FastV, DyCoke, and OmniZip yield lower latency with worse accuracy.
Thus, \textbf{SOC preserves response quality while accelerating inference}.

\subsection{Scaling to 30B Model with More Complex Tasks}
\label{sec:exp_30B}
We scale SOC to Qwen3-Omni-30B-A3B~\cite{xu2025qwen3omni} model for complex long-form generation and multi-turn conversation. 
We report prefill time for LongVALE to avoid influence of response lengths (More in Appendix~\ref{sec:appendix-30b-efficiency}) and total model latency for LongShOT.
Table~\ref{tab:30b-long} highlights two observations:

\textit{i) SOC remains competitive for long-form generation at the 30B scale.}
SOC scores the best on at least one metric across all budgets on LongVALE, extending its capacity to extended generation.
It also substantially reduces prefill time relative to the base model. 
\enlargethispage{1.5\baselineskip}
Faster methods, including OmniZip and random selection, yield lower accuracy, highlighting SOC's favorable quality-efficiency trade-off.

\newpage
\textit{ii) SOC supports complex multi-turn interaction.}
SOC achieves the highest rubric score at every retention tier on LongShOT, preserving effective context for diverse follow-up questions.
At 35\% retention, it delivers a $1.66\times$ speedup over base model versus at most $1.18\times$ for other baselines.
These demonstrate that \textbf{SOC scales effectively to 30B, combining generation quality with notable acceleration}.

\subsection{Hardware Implementation}
\label{sec:method-efficient}
\paragraph{Scoring without attention matrices.} 
Naive saliency scoring materializes the full attention matrix, adding computation and memory overhead.
Our Triton kernel instead computes scores tile by tile, reusing the log-sum-exp statistics returned by FlashAttention-2~\cite{dao2024flashattention2} for softmax normalization.
\begin{wraptable}{r}{0.45\columnwidth}
\vspace{\dimexpr-\baselineskip-0.5pt\relax}
\centering\scriptsize
\captionsetup{font=scriptsize,skip=3pt,justification=raggedright,singlelinecheck=false}
\setlength{\tabcolsep}{1pt}
\makebox[\linewidth][r]{%
\begin{tabular*}{\dimexpr\linewidth+0.03\columnwidth\relax}{@{\extracolsep{\fill}}lrrrr@{}}
module & naive & kernel & gain & $L_1$ \\
\hline
audio & 0.691 & 0.138 & $5.01\times$ & $1.8$ \\
visual & 6.607 & 0.843 & $7.84\times$ & $4.6$ \\
LLM & 28.424 & 2.887 & $9.85\times$ & $17.0$ \\
\hline
\end{tabular*}}
\makebox[\linewidth][r]{%
\begin{minipage}{\dimexpr\linewidth+0.03\columnwidth\relax}
\caption{\textbf{Triton accelerates saliency scoring with minor errors.} Timing in ms; relative $L_1$ in units of $10^{-7}$.}
\label{tab:kernel-implementation}
\end{minipage}}
\end{wraptable}

This avoids storing the full attention matrix and accelerates scoring by $5.01\times$, $7.84\times$, and $9.85\times$ in the audio/visual encoder, and
\WFclear
\noindent LLM as in Table~\ref{tab:kernel-implementation}.
Appendix~\ref{sec:appendix-triton} provides pseudocode.

\paragraph{Integration with vLLM.}
We integrate SOC with vLLM for 30B model. 
Stage I compresses tokens before vLLM request.
Stage II runs inside the engine: a paged-attention controller tracks the effective context lengths for each request and layer.
\begin{wraptable}{r}{0.45\columnwidth}
\vspace{\dimexpr-\baselineskip+8pt\relax}
\captionsetup{font=scriptsize,skip=3pt,justification=raggedright,singlelinecheck=false}
\centering\fontsize{6.5}{7.5}\selectfont
\setlength{\aboverulesep}{0pt}\setlength{\belowrulesep}{0pt}
\setlength{\tabcolsep}{1.2pt}\renewcommand{\arraystretch}{1.05}
\begin{tabularx}{\linewidth}{@{}l *{4}{>{\centering\arraybackslash}X}@{}}
 & \multicolumn{2}{c}{LongShOT} & \multicolumn{2}{c}{LongVALE} \\
\cmidrule(lr{1.5pt}){2-3}\cmidrule(lr{1.5pt}){4-5}
 & base & SOC & base & SOC \\
\hline
HF 1 & 0.199 & \cellcolor{oursrow}-- & 0.091 & \cellcolor{oursrow}-- \\
vLLM 1 & 0.179 & \cellcolor{oursrow}0.310 & 0.077 & \cellcolor{oursrow}0.095 \\
vLLM 4 & 0.238 & \cellcolor{oursrow}0.540 & 0.190 & \cellcolor{oursrow}0.287 \\
vLLM 16 & 0.242 & \cellcolor{oursrow}0.633 & 0.337 & \cellcolor{oursrow}0.452 \\
\hline
\end{tabularx}
\caption{\textbf{SOC improves vLLM throughput across batch sizes.}}
\label{tab:vllm-timing}
\end{wraptable}

Table~\ref{tab:vllm-timing} compares model throughput (turn/s) under vLLM and Hugging Face.
SOC achieves higher vLLM throughput as the batch size increases,
\WFclear
\noindent reaching $2.62\times$/$1.51\times$ speedup on LongShOT/LongVALE.

\subsection{Ablation Studies}
\label{sec:ablation}
We evaluate the contributions of SOC's components and use Qwen2.5-Omni-3B on Daily-Omni unless otherwise stated.

\vspace{-8pt}
\begin{wraptable}{r}{0.55\columnwidth}
\vspace{\dimexpr-\baselineskip-1pt\relax}
\centering\scriptsize
\setlength{\tabcolsep}{0.4pt}\renewcommand{\arraystretch}{1.05}
\captionsetup{font=dfcmixseven,skip=3pt,justification=justified,singlelinecheck=false}
\makebox[\linewidth][r]{%
\begin{minipage}{\dimexpr\linewidth+0.05\columnwidth\relax}
{\fontsize{7}{8}\selectfont
\begin{tabular*}{\linewidth}{@{\extracolsep{\fill}}c*{7}{c}@{}}
$\alpha$ & 0 & 0.1 & 0.3 & 0.5 & \cellcolor{oursrow}0.7 & 0.9 & 1.0 \\
\hline
35\% & 60.4 & 60.6 & 61.1 & 62.2 & \cellcolor{oursrow}\textbf{62.4} & 61.8 & 61.7 \\
45\% & 61.7 & 61.3 & 62.6 & 62.9 & \cellcolor{oursrow}62.9 & \textbf{63.0} & 62.7 \\
\hline
\end{tabular*}}
\captionsetup{width=\linewidth,margin=0pt}
\caption{\textbf{Saliency and DFC work well together.} 
We set $\alpha=0.7$ by default in our experiments.}

\label{tab:ablation-components-b}
\end{minipage}}
\end{wraptable}

\paragraph{Diversity feature coverage.}
Table~\ref{tab:ablation-components-b} varies the saliency-DFC mixing ratio.
\WFclear
\noindent Combining the two outperforms either alone, with $\alpha=0.7$ offering a good trade-off; we adopt it as the default.
We use only Stage I to isolate the effect of the selection rule.

\vspace{-4pt}
\begin{wraptable}{r}{0.45\columnwidth}
\vspace{\dimexpr-\baselineskip-3pt\relax}
\captionsetup{font=dfccompact,skip=3pt,justification=raggedright,singlelinecheck=false}
\noindent\makebox[\linewidth][r]{%
\begin{minipage}{0.49\columnwidth}
\centering\fontsize{6.5}{7.5}\selectfont
\setlength{\tabcolsep}{1pt}
\begin{tabular}{>{\raggedright\arraybackslash}p{\dimexpr\linewidth-34pt\relax}>{\centering\arraybackslash}p{30pt}}
DFC initialization & accuracy \\
\hline
\rowcolor{oursrow}independent & \textbf{62.41} \\
saliency set & 62.24 \\
farthest from saliency & 62.16 \\
\hline
\end{tabular}
\caption{\textbf{Independent DFC initialization remains effective.}}
\label{tab:ablation-fps-init}
\end{minipage}%
}
\end{wraptable}

\paragraph{DFC initialization.}
By default, DFC starts from the farthest token among those remaining after
\WFclear
\noindent stage I.
We compare two alternatives: i) starting with farthest token from saliency selection, and ii) directly using saliency-selected tokens as initialization.
Table~\ref{tab:ablation-fps-init} shows our default independent initialization yields the best result.

\begin{wraptable}{r}{0.45\columnwidth}
\vspace{\dimexpr-\baselineskip+9pt\relax}
\captionsetup{font=dfccompact,skip=3pt,justification=raggedright,singlelinecheck=false}
\noindent\makebox[\linewidth][r]{%
\begin{minipage}{.49\columnwidth}
\centering\fontsize{6.5}{7.5}\selectfont
\setlength{\tabcolsep}{0.7pt}\renewcommand{\arraystretch}{1.05}
\begin{tabular}{>{\raggedright\arraybackslash}p{\dimexpr\linewidth-21.4pt\relax}>{\centering\arraybackslash}p{18.6pt}}
audio selector & acc. \\
\hline
\rowcolor{oursrow}w/ norm. saliency & \textbf{62.4} \\
w/o norm. saliency & 62.2 \\
DFC & 62.0 \\
0.7 w/ norm. saliency + 0.3 DFC  & 62.2 \\
0.7 w/o norm. saliency + 0.3 DFC  & 61.4 \\
adjacent feature change (cosine) &  59.7 \\
\hline
\end{tabular}
\caption{\textbf{Additional heuristics do not improve over audio saliency.}}
\label{tab:ablation-components-c}
\end{minipage}%
}
\end{wraptable}

\paragraph{Audio selection rules.}
Our ablation on audio selection examines its integration with DFC, the role of window-wise normalization, and the effectiveness of adjacent
\WFclear
\noindent feature changes as a selection criterion.
Table~\ref{tab:ablation-components-c} shows that these additional priors again, as anticipated, offer no consistent advantage over simple encoder saliency.

\begin{wraptable}{r}{0.45\columnwidth}
\vspace{\dimexpr-\baselineskip-3pt\relax}
\noindent\makebox[\linewidth][r]{%
\begin{minipage}{0.49\columnwidth}
\centering\scriptsize
\setlength{\tabcolsep}{0.5pt}\renewcommand{\arraystretch}{1.05}

\begin{tabularx}{\linewidth}{>{\raggedright\arraybackslash}p{0.48\linewidth}*{3}{>{\centering\arraybackslash}X}}
\hline
\rowcolor{headerrow}\shortstack[l]{3B, 35\% retention} & T & A+V & all \\
Daily-Omni & \cellcolor{oursrow}\textbf{62.9} & 62.7 & 62.7 \\
WorldSense & \cellcolor{oursrow}\textbf{45.8} & 45.3 & 45.3 \\
Video-MME & \cellcolor{oursrow}\textbf{62.1} & 61.5 & 61.6 \\
AVUT & \cellcolor{oursrow}62.8 & \textbf{63.0} & 62.8 \\
LVOmni & \cellcolor{oursrow}\textbf{33.2} & 32.6 & 32.8 \\
\hline
\rowcolor{headerrow}\shortstack[l]{30B, LongShOT} & T & A+V & all \\
15\% retention & \cellcolor{oursrow}\textbf{22.6} & 21.8 & 21.4 \\
35\% retention & \cellcolor{oursrow}\textbf{28.2} & 27.8 & 27.4 \\
\hline
\end{tabularx}
\caption{\textbf{Text-guided refinement generally exceeds multimodal queries at stage II.}}
\label{tab:ablation-progressive-c}

\end{minipage}%
}
\end{wraptable}

\paragraph{Refinement attention source.} 
Equation~\ref{eq:soc-context-score} aggregates attention from text queries (T), and we further evaluate audio-visual (A+V) queries and all prefill tokens for contextual refinement.
Table~\ref{tab:ablation-progressive-c} shows that text-only guidance matches or exceeds the alternatives,
\WFclear
\noindent with evident gains on LongShOT.
This grasp of task-related information is imperative in multi-turn dialogues, where context need varies across turns.

\subsection{Visualizations}
\label{sec:exp_visualizations}
\suppressfloats[t]
\begin{figure}[t]
\centering
\includegraphics[width=\linewidth]{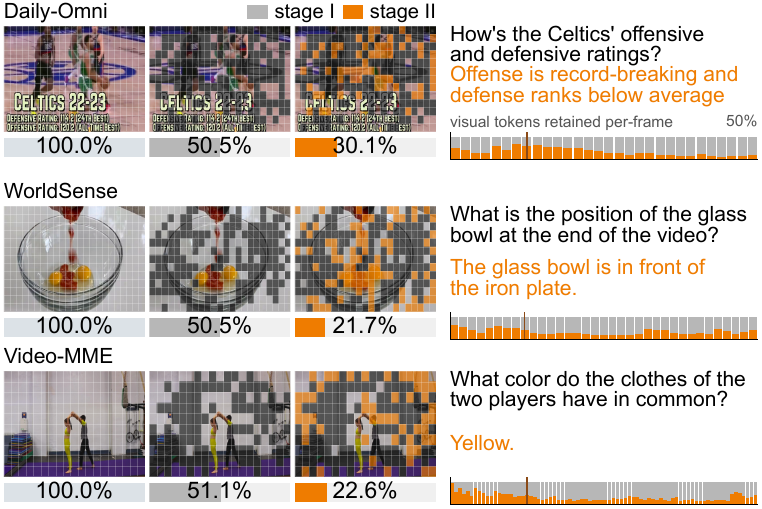}
\caption{SOC's two-stage selection progressively removes visual redundancy while keeping crucial information intact.}
\label{fig:token-visualization}
\end{figure}

Figure~\ref{fig:token-visualization} illustrates SOC's two-stage visual token selection on Daily-Omni, WorldSense, and Video-MME.
It preserves semantically important regions: the statistics of the basketball team, the glass bowl, and colors of the clothes, respectively.
More visualizations are in Appendix~\ref{sec:appendix-visualizations}.

\section{Related Work}
\label{sec:related}

\paragraph{Omni-modal language models.}
Omni-modal language models extend LLMs to multi-modal perception and generation.
OneLLM~\cite{han2023onellm} aligns eight modalities, while NExT-GPT~\cite{wu2024nextgpt} integrates LLM with modality encoders and diffusion decoders.
AnyGPT~\cite{zhan2024anygpt} and MIO~\cite{wang2025mio} unify modalities through discrete autoregressive modeling. 
Qwen2.5-Omni~\cite{Qwen2.5-Omni} introduces a Thinker-Talker architecture, and Qwen3-Omni~\cite{xu2025qwen3omni} scales it with mixture-of-experts.
OmniVinci~\cite{ye2026omnivinci} improves audio-visual alignment and temporal modeling, while NExT-OMNI~\cite{luo2026nextomni} unifies understanding and generation via discrete flow matching.
These advances in omni-modal capabilities also motivate efficient solutions of increasingly long multi-modal contexts.

\paragraph{Omni-modal token compression.}
\enlargethispage{0pt}
\begingroup
\brokenpenalty=0
Compressing multi-modal context accelerates OLLM inference.
FrameFusion~\cite{fu2025framefusion} and AIM~\cite{zhong2025aim} merge similar visual tokens, while DyCoke~\cite{tao2025dycoke} additionally employs KV cache compression.
AKS~\cite{tang2025aks} and LongVU~\cite{shen2025longvu} prioritize selected frames based on estimated importance.
FastV~\cite{chen2024fastv} prunes visual tokens using model attention, while VisionZip~\cite{yang2025visionzip} employs encoder attention.
Prune2Drive~\cite{xiong2026prune2drive} and DivPrune~\cite{alvar2025divprune} remove visual tokens using diversity.
OmniZip~\cite{tao2026omnizip} guides compression with audio. 
OmniPrune~\cite{zhong2026omniprune}, OmniPack~\cite{su2026omnipack}, and OmniFit~\cite{wang2026omnifit} use token-pairing, while OmniSelect~\cite{yang2026omniselect} relies on external model guidance. 
These overheads can limit speedups.
SOC avoids complex recomputation and yields excellent performance with practical acceleration.
\par\endgroup

\section{Conclusion}
\label{sec:conclusion}
This paper introduces Simple Omni Compression (SOC), a method that eliminates complex inductive biases and focuses on retaining crucial information with minimal extra overhead. 
SOC uses each modality's own encoder to select tokens, distributes visual budget evenly over time, while ensuring feature diversity coverage. 
Once multimodal representations are sufficiently contextualized, it further refines the visual context via middle-layer attention from question tokens.
Extensive experiments on omni-modal QA, long-form generation, and multi-turn dialogue demonstrate SOC's capacity across model scales and retention ratios.
By avoiding elaborate heuristics and costly computations, it translates token reduction into practical inference speedups.
Our work encourages a first-principles approach to omni-modal compression: identify which information matters before imposing structural assumptions. 
We hope this perspective inspires future research.

\enlargethispage{\baselineskip}
\section*{Acknowledgement}
We appreciate the valuable discussion, support and inspiration from Jiaxin Wu, Junhao Huang, Haoyang Chen, and Jingle Fu throughout this work.

\section*{LLM Usage Statement}
We use OpenAI Codex to assist with polishing the manuscript's writing, debugging code implementation, and improving the aesthetics of figures and diagrams.
All scientific contributions, experimental results, and conclusions were produced and verified by the authors, who take full responsibility for the content of this paper.

{
    \small
    \bibliographystyle{ieeenat_fullname}
    \bibliography{main}
}


\clearpage
\appendix

\noindent We organize our appendix as follows:

\noindent\textbf{Evaluation and Implementation Details}

\noindent Section~\ref{sec:appendix-details}: Evaluation and Implementation Details.

\noindent Section~\ref{sec:appendix-triton}: Triton Pseudocode.

\noindent\textbf{Additional Experimental Results}

\noindent Section~\ref{sec:apx-complete-qa-results}: Complete 3B/7B experiment results.

\noindent Section~\ref{sec:appendix-efficiency}: 3B/7B efficiency measurements.

\noindent Section~\ref{sec:appendix-30b-efficiency}: 30B experiments' efficiency analyses.

\noindent Section~\ref{sec:appendix-visualizations}: More visualizations.

\noindent\textbf{Additional Analyses and Ablation Studies}

\noindent Section~\ref{sec:appendix-attention-analysis}: Temporal attention pattern analyses.

\noindent Section~\ref{sec:appendix-alternatives}: Alternative compression designs.

\noindent Section~\ref{sec:appendix-more-ablations}: More ablation studies.

\section{Evaluation and Implementation Details}
\label{sec:appendix-details}

\subsection{Benchmarks}
\label{sec:appendix-benchmarks}
\noindent\textbf{Daily-Omni}~\cite{zhou2025dailyomni} evaluates audio-visual reasoning in everyday videos, with questions regarding from both modalities. 
It contains 1,197 multiple-choice questions over 684 videos.

\noindent\textbf{WorldSense}~\cite{hong2026worldsense} evaluates real-world omni-modal perception and reasoning across diverse domains. 
It has 3,172 multiple-choice questions over 1,662 videos.

\noindent\textbf{Video-MME}~\cite{fu2025videomme} measures video understanding across videos with various durations. 
The complete split contains 2,700 questions over 900 videos.

\noindent\textbf{AVUT}~\cite{yang2025avut} focuses on audio-centric video understanding and is designed to reduce text shortcuts. 
The complete evaluation split contains 1,734 questions over 691 videos.

\noindent\textbf{LVOmniBench}~\cite{tao2026lvomnibench} evaluates long audio-video understanding, emphasizing information integration over extended context. 
It features 1,014 questions over 275 videos.

\noindent\textbf{LongVALE}~\cite{geng2025longvale} requires model to generate temporally localized event annotations for long videos with visual and audio evidence. 
Its test split contains 1,171 videos.

\noindent\textbf{LongShOT}~\cite{kurpath2025longshot} evaluates open-ended reasoning in long videos through multi-turn dialogue. 
Evaluation incorporates 3,272 scored samples from 262 videos.

\subsection{Baselines}
\label{sec:appendix-baselines}
\noindent\textbf{DyCoke.}
DyCoke~\cite{tao2025dycoke} exploits visual redundancy through token merging and KV cache compression. 

\noindent\textbf{FastV.}
FastV~\cite{chen2024fastv} uses attention from early LLM layers to identify and discard visual tokens in subsequent layers.

\noindent\textbf{SparseVLM.}
SparseVLM~\cite{zhang2025sparsevlm} progressively sparsifies visual tokens using text-guided attention scores and recycles information from discarded tokens.

\noindent\textbf{OmniPrune.}
OmniPrune~\cite{zhong2026omniprune} scores multimodal tokens using encoder attention and query relevance, with audio-guided temporal windows and audio-video pairing.

\noindent\textbf{OmniPack.}
OmniPack~\cite{su2026omnipack} combines importance selection, DPC-KNN coverage, and token merging before the LLM, followed by further compression within the LLM.

\noindent\textbf{OmniZip.}
OmniZip~\cite{tao2026omnizip} uses audio-guided importance to allocate visual compression dynamically across temporal groups while also reducing audio tokens.

\subsection{Token Budgets and Scoring Protocols}

\noindent\textbf{Token budgets.}
We match all methods by layer-average multimodal token retention, $\bar r=\frac{1}{L}\sum_{l=0}^{L-1}r^{(l)}$.
With middle layer refinement, $\bar r=(r_1+r_2)/2$.
For $\bar r=15\%/20\%/35\%/45\%$, we use $(r_1,r_2)=(20\%,10\%)/(30\%,10\%)/(50\%,20\%)/(50\%,40\%)$ and retain $35\%/35\%/45\%/55\%$ of audio tokens, allocating the rest to visual tokens.
Appendices~\ref{sec:appendix-refinement-layer} and~\ref{sec:appendix-schedules} evaluate alternative refinement locations and budget splits.

\noindent\textbf{LongShOT and multi-turn refinement.}
Following the official implementation, we generate each target turn using reference-answer history, 32 uniformly sampled video frames, and full audio.
As audio accounts for 78\% of input tokens, SOC retains $20\%/35\%/40\%/45\%$ of visual tokens for four retentions and allocates the rest to audio.
Stage I compression is shared across turns, while stage II refines audio for each turn using textual question's queries.

\section{Triton Saliency Scoring}
\label{sec:appendix-triton}
Algorithm~\ref{alg:triton-saliency} illustrates how our Triton kernel computes attention saliency by aggregating attention over selected queries $\mathcal{I}$.
For block-wise encoder attention, $\mathcal{I}$ means attention block as in Equation~\ref{eq:soc-encoder-score}; for contextual refinement in Equation~\ref{eq:soc-context-score}, it contains all textual question queries.
The kernel computes QK products in tiles, accumulates partial saliency scores in FP32, and reuses the row-wise log-sum-exp from FlashAttention-2 for softmax normalization.

\begin{algorithm}[!ht]
\caption{Triton saliency scoring kernel}
\label{alg:triton-saliency}
\footnotesize
\begin{algorithmic}[1]
\Require single-headed queries $Q$, keys $K$, row log-sum-exp $L$, selected queries $\mathcal{I}$, attention mask $\mathcal{M}$ (causal, block, and padding constraints).
\State Divide $Q \in \mathbb{R}^{|\mathcal{I}|\times d}$ into $T_q=\lceil |\mathcal{I}|/B_q\rceil$ tiles of size $B_q\times d$.
\State Divide $K \in \mathbb{R}^{N\times d}$ into $T_k=\lceil N/B_k\rceil$ tiles of size $B_k\times d$.
\For{$1 \leq j \leq T_k}$ in parallel
    \State Load $K_j$ into on-chip memory
    \State $a_j\gets\mathbf{0}$
    \For{$1 \leq i \leq T_q}$
        \State Load $Q_i$ and $L_i$
        \State $Z\gets Q_{i}K_{j}^{\top}/\sqrt{d}$
        \State $A\gets\exp(Z+\mathcal{M}_{ij}-L_i\mathbf{1}^{\top})$
        \State $a_j\gets a_j+\operatorname{colsum}(A)$
    \EndFor
    \State Write $R_j\gets a_j$ to global memory
\EndFor
\State \Return single-head scores $R$
\end{algorithmic}
\end{algorithm}

The kernel runs in parallel across heads, for GQA, head $h$ uses KV head $\lfloor h/(H/H_{\mathrm{KV}})\rfloor$. 
A separate kernel normalizes scores as $S=\sum_{h=0}^{H-1}R^{(h)}/D$, where $D=H$ for encoder saliency and $D=H|\mathcal{I}|$ for contextual refinement.
We use $B_q = 32$ and $B_k = 64$ in our implementation.

\onecolumn
\section{Complete Question-Answering Results}
\label{sec:apx-complete-qa-results}
Table~\ref{tab:appendix-performance-paired} presents results at all four retention tiers (15\%, 20\%, 35\%, and 45\%) on the 3B/7B model. 
SOC keeps 35\%, 35\%, 45\%, and 55\% of audio tokens for multimodal token budget of 15\%, 20\%, 35\%, and 45\%, respectively.

\begin{table}[H]
\centering\scriptsize
\setlength{\tabcolsep}{2pt}\renewcommand{\arraystretch}{1.05}
\begin{tabularx}{\textwidth}{l*{10}{>{\centering\arraybackslash}X}}
 & \multicolumn{5}{c}{\normalsize Qwen2.5-Omni-3B} & \multicolumn{5}{c}{\normalsize Qwen2.5-Omni-7B} \\
\cmidrule(lr){2-6}\cmidrule(lr){7-11}
method & Daily-Omni & WorldSense & Video-MME & AVUT & {\fontsize{6}{7}\selectfont\mbox{LVOmniBench}} & Daily-Omni & WorldSense & Video-MME & AVUT & {\fontsize{6}{7}\selectfont\mbox{LVOmniBench}} \\
\hline
\rowcolor{fullrow}Full & 62.5 & 46.4 & 62.4 & 63.4 & 33.0 & 63.9 & 46.6 & 64.8 & 65.6 & 32.5 \\
\hline\rowcolor{headerrow}\multicolumn{11}{l}{\textbf{45\%}} \\
random & 58.7 & 43.7 & 60.7 & 60.6 & 31.5 & 59.4 & 44.4 & 63.5 & 62.1 & 31.3 \\
DyCoke & 60.6 & \textbf{46.0} & 60.5 & 62.5 & 31.3 & 61.8 & \underline{46.5} & 62.7 & 64.6 & 30.9 \\
FastV & 62.2 & 45.7 & 61.3 & 61.7 & 31.9 & 63.3 & 45.0 & 63.1 & 64.6 & \underline{32.4} \\
SparseVLM & 62.1 & 45.5 & 61.4 & 62.4 & 31.6 & \textbf{63.7} & 45.8 & \underline{64.1} & \underline{64.8} & 31.0 \\
OmniPrune & \underline{63.0} & 45.6 & \textbf{62.1} & \textbf{63.4} & \textbf{32.9} & \underline{63.6} & 45.7 & 63.6 & \textbf{65.3} & \textbf{32.5} \\
OmniPack & 60.6 & \underline{45.9} & 61.6 & 62.3 & 31.9 & 62.7 & 46.2 & \textbf{64.2} & 64.1 & 31.5 \\
OmniZip & 60.8 & \underline{45.9} & 60.0 & 61.9 & 31.4 & 62.8 & 45.9 & 62.7 & 64.3 & 31.5 \\
\rowcolor{oursrow}SOC (Ours) & \textbf{63.4} & 45.5 & \underline{61.9} & \underline{63.1} & \underline{32.8} & 62.7 & \textbf{46.6} & \underline{64.1} & 64.4 & \textbf{32.5} \\
\hline\rowcolor{headerrow}\multicolumn{11}{l}{\textbf{35\%}} \\
random & 56.8 & 42.8 & 59.5 & 59.0 & 31.6 & 56.7 & 43.2 & 62.7 & 60.4 & 31.0 \\
DyCoke & 59.2 & 45.1 & 58.9 & 61.8 & 30.7 & 59.9 & 45.4 & 60.3 & \textbf{64.9} & 29.4 \\
FastV & 60.9 & 45.2 & 59.5 & 59.6 & 31.2 & 61.8 & 44.2 & 61.1 & 63.0 & 31.7 \\
SparseVLM & \underline{62.5} & 45.2 & 61.3 & 62.1 & 32.1 & 62.2 & 45.3 & 63.1 & 64.5 & \textbf{32.1} \\
OmniPrune & 61.7 & 45.4 & \underline{61.8} & \underline{62.6} & \underline{32.8} & \textbf{62.9} & 45.5 & 63.5 & 64.6 & 31.8 \\
OmniPack & 59.6 & \underline{45.5} & 61.1 & 61.5 & \underline{32.8} & \underline{62.4} & \underline{45.6} & \underline{63.7} & 63.2 & \underline{31.9} \\
OmniZip & 59.5 & \textbf{45.8} & 58.7 & 61.4 & 30.9 & 61.8 & \textbf{45.7} & 60.2 & 64.5 & 30.8 \\
\rowcolor{oursrow}SOC (Ours) & \textbf{62.9} & \textbf{45.8} & \textbf{62.1} & \textbf{62.8} & \textbf{33.2} & 62.2 & \textbf{45.7} & \textbf{64.9} & \underline{64.8} & \textbf{32.1} \\
\hline\rowcolor{headerrow}\multicolumn{11}{l}{\textbf{20\%}} \\
random & 50.8 & 40.4 & 57.0 & 54.7 & 31.5 & 50.9 & 40.4 & 59.4 & 56.7 & 30.8 \\
DyCoke & 57.3 & \underline{44.6} & N/A & 58.7 & N/A & 58.0 & 43.0 & N/A & 61.0 & N/A \\
FastV & 55.6 & 42.8 & 55.3 & 54.6 & \underline{32.6} & 56.7 & 41.9 & 51.4 & 56.9 & 30.9 \\
SparseVLM & 58.6 & 42.4 & 58.4 & 56.6 & 32.5 & \underline{59.6} & 43.2 & 60.3 & 60.9 & 31.1 \\
OmniPrune & \underline{58.9} & 44.1 & \underline{59.3} & 60.5 & \underline{32.6} & \underline{59.6} & \textbf{44.5} & \underline{62.9} & \textbf{63.8} & \underline{31.6} \\
OmniPack & 55.6 & 43.5 & 59.2 & 58.5 & 32.1 & 57.1 & 43.2 & 62.3 & 61.9 & 30.6 \\
OmniZip & 57.6 & 44.4 & 56.0 & \underline{60.8} & 31.7 & \underline{59.6} & 44.2 & 58.6 & 62.2 & 30.1 \\
\rowcolor{oursrow}SOC (Ours) & \textbf{60.9} & \textbf{44.7} & \textbf{60.6} & \textbf{62.1} & \textbf{33.8} & \textbf{60.1} & \underline{44.4} & \textbf{63.0} & \underline{62.5} & \textbf{31.7} \\
\hline\rowcolor{headerrow}\multicolumn{11}{l}{\textbf{15\%}} \\
random & 48.5 & 39.7 & 55.7 & 53.2 & 31.6 & 48.0 & 39.0 & 58.5 & 54.5 & 30.5 \\
DyCoke & 53.9 & N/A & N/A & 56.2 & N/A & 54.7 & N/A & N/A & 60.2 & N/A \\
FastV & 51.4 & 42.2 & 53.0 & 53.5 & 32.2 & 54.5 & 39.5 & 52.0 & 53.4 & 30.3 \\
SparseVLM & 55.6 & 41.7 & 56.1 & 54.6 & \textbf{33.4} & 57.6 & \underline{43.2} & 56.9 & 57.9 & \textbf{31.3} \\
OmniPrune & \underline{56.9} & 42.5 & 57.0 & \underline{59.1} & \underline{33.2} & \underline{58.9} & 42.2 & 60.6 & \underline{61.1} & \underline{31.0} \\
OmniPack & 52.1 & 42.1 & \underline{58.4} & 57.0 & 31.6 & 55.2 & 41.0 & \underline{61.5} & 59.2 & 30.8 \\
OmniZip & 54.7 & \underline{43.7} & N/A & 58.0 & 31.7 & 56.9 & 43.1 & N/A & 60.6 & 29.3 \\
\rowcolor{oursrow}SOC (Ours) & \textbf{59.1} & \textbf{43.9} & \textbf{58.9} & \textbf{61.1} & \textbf{33.4} & \textbf{59.1} & \textbf{43.7} & \textbf{62.0} & \textbf{61.6} & 30.6 \\
\hline\end{tabularx}
\caption{Omni-modal QA benchmarks results across all four retention tiers for Qwen2.5-Omni-3B/7B.}
\label{tab:appendix-performance-paired}\label{tab:appendix-performance-3b}\label{tab:appendix-performance-7b}
\end{table}

\clearpage
\section{Detailed 3B/7B Efficiency Measurements}
\label{sec:appendix-efficiency}
Table~\ref{tab:appendix-efficiency-full-3b} reports model latency and FLOPs for the configurations in the 3B/7B performance table.
Each benchmark's profiling is evaluated on the same 100 questions.
Model latency includes encoding, token selection/compression, prefill, and decoding.
\begin{table}[H]
\centering\scriptsize
\captionsetup{font=scriptsize}
\setlength{\tabcolsep}{3pt}\renewcommand{\arraystretch}{0.88}
\begin{tabularx}{\textwidth}{l*{10}{>{\centering\arraybackslash}X}}

\multirow{2}{*}{method} & \multicolumn{2}{c}{Daily-Omni} & \multicolumn{2}{c}{WorldSense} & \multicolumn{2}{c}{Video-MME} & \multicolumn{2}{c}{AVUT} & \multicolumn{2}{c}{LVOmniBench} \\
 & time (s) & TFLOPs & time (s) & TFLOPs & time (s) & TFLOPs & time (s) & TFLOPs & time (s) & TFLOPs \\
\hline
\rowcolor{fullrow}3B-base & 0.712 & 160.6 & 0.990 & 241.4 & 1.151 & 300.2 & 0.915 & 215.6 & 1.263 & 336.2 \\
\rowcolor{headerrow}\multicolumn{11}{l}{\textbf{45\%}} \\
random & 0.574 & 100.5 & 0.765 & 143.8 & 0.849 & 171.6 & 0.717 & 131.8 & 0.912 & 188.3 \\
DyCoke & 0.573 & 100.1 & 0.778 & 145.6 & 0.860 & 173.9 & 0.723 & 132.2 & 0.910 & 187.3 \\
FastV & 0.566 & 101.0 & 0.756 & 144.7 & 0.844 & 173.0 & 0.709 & 132.5 & 0.906 & 189.9 \\
SparseVLM & 0.740 & 101.3 & 1.033 & 145.4 & 1.153 & 174.0 & 0.936 & 133.1 & 1.246 & 191.2 \\
OmniPrune & 1.000 & 100.6 & 1.445 & 144.0 & 1.615 & 171.7 & 1.273 & 132.2 & 1.781 & 188.8 \\
OmniPack & 0.848 & 101.0 & 1.196 & 144.6 & 1.384 & 172.8 & 1.099 & 132.6 & 1.528 & 189.7 \\
OmniZip & 0.603 & 101.0 & 0.831 & 146.2 & 0.951 & 178.8 & 0.760 & 133.4 & 1.020 & 190.2 \\
\rowcolor{oursrow}SOC (Ours) & 0.716 & 101.4 & 0.977 & 144.6 & 1.014 & 171.8 & 0.856 & 132.0 & 1.186 & 190.9 \\
\hline\rowcolor{headerrow}\multicolumn{11}{l}{\textbf{35\%}} \\
random & 0.553 & 91.3 & 0.734 & 129.4 & 0.804 & 153.2 & 0.688 & 119.3 & 0.859 & 167.3 \\
DyCoke & 0.555 & 90.9 & 0.745 & 130.3 & 0.818 & 155.5 & 0.693 & 119.7 & 0.859 & 167.3 \\
FastV & 0.541 & 92.0 & 0.725 & 130.7 & 0.801 & 155.0 & 0.683 & 120.3 & 0.858 & 169.6 \\
SparseVLM & 0.692 & 92.4 & 0.959 & 131.4 & 1.061 & 156.2 & 0.876 & 120.7 & 1.161 & 170.9 \\
OmniPrune & 0.992 & 91.8 & 1.451 & 129.3 & 1.606 & 153.3 & 1.269 & 119.3 & 1.785 & 167.3 \\
OmniPack & 0.759 & 91.7 & 1.047 & 128.7 & 1.217 & 153.9 & 0.976 & 119.8 & 1.320 & 166.4 \\
OmniZip & 0.580 & 91.2 & 0.799 & 132.4 & 0.909 & 161.8 & 0.733 & 121.1 & 0.967 & 169.3 \\
\rowcolor{oursrow}SOC (Ours) & 0.676 & 92.1 & 0.913 & 130.9 & 1.025 & 155.2 & 0.848 & 120.5 & 1.005 & 167.4 \\
\hline\rowcolor{headerrow}\multicolumn{11}{l}{\textbf{20\%}} \\
random & 0.525 & 78.5 & 0.691 & 109.7 & 0.750 & 128.3 & 0.651 & 102.0 & 0.793 & 139.2 \\
DyCoke & 0.535 & 78.3 & 0.716 & 111.6 & N/A & N/A & 0.662 & 102.7 & N/A & N/A \\
FastV & 0.520 & 79.2 & 0.688 & 111.3 & 0.751 & 131.1 & 0.644 & 103.6 & 0.794 & 142.6 \\
SparseVLM & 0.618 & 79.8 & 0.834 & 112.3 & 0.919 & 132.0 & 0.770 & 104.1 & 0.989 & 143.7 \\
OmniPrune & 1.000 & 78.7 & 1.467 & 110.1 & 1.616 & 128.7 & 1.276 & 102.3 & 1.793 & 138.4 \\
OmniPack & 0.645 & 78.7 & 0.882 & 110.0 & 0.980 & 128.6 & 0.816 & 101.8 & 1.060 & 139.5 \\
OmniZip & 0.551 & 78.6 & 0.759 & 112.8 & 0.851 & 133.4 & 0.695 & 104.4 & 0.905 & 140.0 \\
\rowcolor{oursrow}SOC (Ours) & 0.599 & 78.9 & 0.773 & 109.9 & 0.903 & 129.2 & 0.756 & 102.7 & 0.967 & 140.3 \\
\hline\rowcolor{headerrow}\multicolumn{11}{l}{\textbf{15\%}} \\
random & 0.522 & 74.5 & 0.677 & 103.7 & 0.733 & 120.7 & 0.645 & 96.7 & 0.774 & 130.7 \\
DyCoke & 0.531 & 74.4 & N/A & N/A & N/A & N/A & 0.655 & 97.4 & N/A & N/A \\
FastV & 0.512 & 75.6 & 0.681 & 106.2 & 0.736 & 123.9 & 0.636 & 98.4 & 0.778 & 134.5 \\
SparseVLM & 0.591 & 75.9 & 0.790 & 106.4 & 0.865 & 124.6 & 0.730 & 98.5 & 0.925 & 135.3 \\
OmniPrune & 1.004 & 74.6 & 1.471 & 103.9 & 1.628 & 120.7 & 1.291 & 96.7 & 1.783 & 130.8 \\
OmniPack & 0.614 & 74.7 & 0.836 & 103.9 & 0.911 & 121.0 & 0.772 & 96.9 & 0.975 & 130.9 \\
OmniZip & 0.552 & 74.4 & 0.752 & 106.2 & N/A & N/A & 0.681 & 98.0 & 0.889 & 132.2 \\
\rowcolor{oursrow}SOC (Ours) & 0.570 & 74.7 & 0.766 & 104.0 & 0.843 & 121.1 & 0.720 & 97.0 & 0.901 & 131.1 \\
\hline
\rowcolor{fullrow}7B-base & 0.899 & 266.3 & 1.266 & 398.9 & 1.513 & 495.9 & 1.155 & 355.5 & 1.657 & 555.4 \\
\rowcolor{headerrow}\multicolumn{11}{l}{\textbf{45\%}} \\
random & 0.673 & 146.3 & 0.883 & 210.5 & 0.996 & 253.4 & 0.817 & 191.5 & 1.061 & 279.4 \\
DyCoke & 0.683 & 145.1 & 0.900 & 213.4 & 1.013 & 257.4 & 0.829 & 192.3 & 1.066 & 279.2 \\
FastV & 0.661 & 147.1 & 0.877 & 212.1 & 0.997 & 255.8 & 0.813 & 192.8 & 1.061 & 282.2 \\
SparseVLM & 0.833 & 147.7 & 1.117 & 213.2 & 1.295 & 257.4 & 1.028 & 193.7 & 1.391 & 284.1 \\
OmniPrune & 1.091 & 146.4 & 1.502 & 211.4 & 1.765 & 253.4 & 1.376 & 192.2 & 1.909 & 279.4 \\
OmniPack & 1.008 & 147.0 & 1.433 & 211.8 & 1.702 & 255.3 & 1.305 & 192.7 & 1.863 & 281.6 \\
OmniZip & 0.709 & 147.1 & 0.951 & 214.5 & 1.111 & 266.0 & 0.862 & 195.7 & 1.174 & 282.9 \\
\rowcolor{oursrow}SOC (Ours) & 0.775 & 146.4 & 1.060 & 211.1 & 1.239 & 256.5 & 1.044 & 195.8 & 1.269 & 280.3 \\
\hline\rowcolor{headerrow}\multicolumn{11}{l}{\textbf{35\%}} \\
random & 0.638 & 126.8 & 0.826 & 180.8 & 0.921 & 216.0 & 0.768 & 165.4 & 0.976 & 237.2 \\
DyCoke & 0.646 & 125.9 & 0.839 & 182.4 & 0.937 & 220.3 & 0.785 & 167.1 & 0.978 & 237.0 \\
FastV & 0.625 & 128.0 & 0.820 & 183.1 & 0.926 & 219.4 & 0.763 & 167.2 & 0.980 & 241.2 \\
SparseVLM & 0.766 & 128.6 & 1.027 & 185.3 & 1.178 & 221.1 & 0.946 & 168.2 & 1.269 & 244.6 \\
OmniPrune & 1.083 & 127.8 & 1.477 & 180.5 & 1.736 & 216.0 & 1.343 & 166.8 & 1.876 & 237.3 \\
OmniPack & 0.881 & 127.3 & 1.229 & 181.6 & 1.448 & 217.2 & 1.116 & 166.2 & 1.582 & 238.5 \\
OmniZip & 0.668 & 126.4 & 0.892 & 185.7 & 1.044 & 231.9 & 0.813 & 170.4 & 1.088 & 241.0 \\
\rowcolor{oursrow}SOC (Ours) & 0.746 & 127.9 & 1.008 & 182.8 & 1.146 & 218.9 & 0.934 & 167.1 & 1.123 & 237.3 \\
\hline\rowcolor{headerrow}\multicolumn{11}{l}{\textbf{20\%}} \\
random & 0.591 & 99.0 & 0.752 & 138.9 & 0.821 & 163.8 & 0.699 & 128.3 & 0.860 & 178.5 \\
DyCoke & 0.597 & 98.6 & 0.761 & 142.5 & N/A & N/A & 0.712 & 129.6 & N/A & N/A \\
FastV & 0.575 & 100.7 & 0.751 & 145.0 & 0.838 & 172.0 & 0.698 & 133.4 & 0.873 & 184.5 \\
SparseVLM & 0.662 & 101.2 & 0.857 & 143.2 & 0.970 & 170.0 & 0.805 & 131.7 & 1.031 & 185.9 \\
OmniPrune & 1.071 & 99.1 & 1.450 & 139.6 & 1.705 & 164.4 & 1.331 & 128.6 & 1.824 & 176.7 \\
OmniPack & 0.731 & 99.2 & 0.967 & 139.3 & 1.099 & 164.2 & 0.892 & 128.7 & 1.177 & 179.0 \\
OmniZip & 0.615 & 99.2 & 0.814 & 143.4 & 0.936 & 173.2 & 0.739 & 131.5 & 0.978 & 179.9 \\
\rowcolor{oursrow}SOC (Ours) & 0.655 & 99.5 & 0.869 & 139.9 & 0.975 & 165.1 & 0.805 & 129.1 & 0.953 & 178.6 \\
\hline\rowcolor{headerrow}\multicolumn{11}{l}{\textbf{15\%}} \\
random & 0.574 & 90.0 & 0.726 & 125.7 & 0.789 & 147.4 & 0.679 & 116.5 & 0.827 & 160.1 \\
DyCoke & 0.579 & 89.9 & N/A & N/A & N/A & N/A & 0.694 & 117.9 & N/A & N/A \\
FastV & 0.572 & 92.1 & 0.728 & 129.6 & 0.805 & 153.1 & 0.683 & 119.7 & 0.841 & 167.0 \\
SparseVLM & 0.623 & 92.3 & 0.800 & 130.1 & 0.895 & 153.8 & 0.756 & 120.1 & 0.945 & 167.9 \\
OmniPrune & 1.069 & 90.1 & 1.449 & 125.9 & 1.687 & 147.2 & 1.325 & 116.4 & 1.815 & 160.3 \\
OmniPack & 0.681 & 90.2 & 0.894 & 126.0 & 1.001 & 147.7 & 0.828 & 116.8 & 1.067 & 160.5 \\
OmniZip & 0.600 & 88.6 & 0.790 & 129.8 & N/A & N/A & 0.728 & 118.7 & 0.947 & 163.1 \\
\rowcolor{oursrow}SOC (Ours) & 0.618 & 90.3 & 0.814 & 126.0 & 0.903 & 147.8 & 0.752 & 116.9 & 0.911 & 160.2 \\
\hline

\end{tabularx}
\caption{\textbf{Complete 3B/7B efficiency measurements.} Model latency (s) and TFLOPs for the configurations in Table~\ref{tab:appendix-performance-paired}, measured on 100 fixed inputs per benchmark.}
\label{tab:appendix-efficiency-full-3b}\label{tab:appendix-efficiency-full-7b}
\end{table}

\clearpage

\section{Complex Task Efficiency Analyses}
\label{sec:appendix-30b-efficiency}
Table~\ref{tab:appendix-longvale-efficiency-complete} decomposes LongVALE inference latency into embedding, selection/compression, LLM prefill, and decoding over the same 100 fixed videos. 
To exclude the influence of varied response length, Table~\ref{tab:30b-long} reports the combined embedding, compression, and LLM prefill time for LongVALE, and total model latency for LongShOT.

\begin{table}[H]
\centering\scriptsize
\setlength{\tabcolsep}{2.5pt}\renewcommand{\arraystretch}{0.9}
\begin{tabularx}{\textwidth}{>{\raggedright\arraybackslash}p{0.30\textwidth}*{6}{>{\hsize=.9\hsize\linewidth=\hsize\centering\arraybackslash}X}>{\hsize=1.7\hsize\linewidth=\hsize\centering\arraybackslash}X>{\hsize=.9\hsize\linewidth=\hsize\centering\arraybackslash}X}
 & TFLOPs & emb. & select. & prefill & decode & model & \mbox{\# output tokens} & \mbox{peak mem.}  \\
\hline
\rowcolor{fullrow}LongVALE on full & 308.8 & 0.460 & 0.248 & 0.841 & 10.735 & 12.284 & 287.4 & 114.22  \\
\rowcolor{headerrow}\multicolumn{9}{l}{\textbf{45\%}} \\
OmniPrune & 137.6 & 0.452 & 0.801 & 0.343 & 9.510 & 11.107 & 255.6 & 114.17  \\
OmniPack & 139.6 & 0.460 & 0.183 & 0.851 & 7.555 & 9.049 & 204.9 & 114.96  \\
OmniZip & 138.5 & 0.463 & 0.186 & 0.322 & 8.335 & 9.305 & 223.5 & 114.24  \\
random (mean) & 137.0 & 0.453 & 0.116 & 0.333 & 8.551 & 9.453 & 229.5 & 114.17  \\
\rowcolor{oursrow}SOC (Ours) & 138.1 & 0.451 & 0.344 & 0.358 & 9.317 & 10.470 & 246.9 & 114.25  \\
\hline\rowcolor{headerrow}\multicolumn{9}{l}{\textbf{35\%}} \\
OmniPrune & 114.7 & 0.462 & 0.825 & 0.266 & 7.271 & 8.823 & 197.7 & 114.19  \\
OmniPack & 117.2 & 0.461 & 0.145 & 0.613 & 5.759 & 6.977 & 154.1 & 115.33  \\
OmniZip & 117.4 & 0.454 & 0.158 & 0.272 & 8.557 & 9.440 & 227.7 & 114.18  \\
random (mean) & 115.3 & 0.456 & 0.097 & 0.250 & 7.990 & 8.793 & 217.9 & 114.19  \\
\rowcolor{oursrow}SOC (Ours) & 119.2 & 0.453 & 0.356 & 0.289 & 8.898 & 9.997 & 237.6 & 114.25  \\
\hline\rowcolor{headerrow}\multicolumn{9}{l}{\textbf{20\%}} \\
OmniPrune & 88.5 & 0.453 & 0.849 & 0.158 & 7.075 & 8.535 & 193.6 & 114.21  \\
OmniPack & 88.5 & 0.452 & 0.118 & 0.338 & 6.929 & 7.836 & 182.7 & 115.78  \\
OmniZip & 90.4 & 0.461 & 0.142 & 0.165 & 7.313 & 8.080 & 199.2 & 114.23  \\
random (mean) & 88.2 & 0.459 & 0.080 & 0.152 & 7.362 & 8.052 & 200.4 & 114.20  \\
\rowcolor{oursrow}SOC (Ours) & 89.7 & 0.452 & 0.245 & 0.167 & 6.931 & 7.794 & 186.4 & 114.17  \\
\hline\rowcolor{headerrow}\multicolumn{9}{l}{\textbf{15\%}} \\
OmniPrune & 80.5 & 0.454 & 0.873 & 0.123 & 5.635 & 7.084 & 153.4 & 114.23  \\
OmniPack & 80.8 & 0.451 & 0.120 & 0.259 & 5.462 & 6.292 & 149.9 & 115.97  \\
OmniZip & 82.3 & 0.457 & 0.139 & 0.129 & 6.436 & 7.161 & 175.2 & 114.25  \\
random (mean) & 80.7 & 0.457 & 0.069 & 0.115 & 7.273 & 7.914 & 197.3 & 114.22  \\
\rowcolor{oursrow}SOC (Ours) & 81.0 & 0.453 & 0.197 & 0.129 & 6.545 & 7.325 & 174.4 & 114.21  \\
\hline
\rowcolor{fullrow}LongShOT on full & 1009.9 & 0.362 & 0.296 & 2.404 & 1.888 & 4.950 & 52.8 & 116.56 \\
\rowcolor{headerrow}\multicolumn{9}{l}{\textbf{45\%}} \\
OmniPrune & 323.7 & 0.362 & 1.808 & 0.790 & 1.821 & 4.782 & 50.7 & 115.00  \\
OmniPack & 338.6 & 0.357 & 0.421 & 2.309 & 1.953 & 5.039 & 54.0 & 119.53  \\
random (mean) & 324.7 & 0.361 & 0.133 & 0.862 & 2.426 & 3.783 & 66.1 & 115.00  \\
\rowcolor{oursrow}SOC (Ours) & 327.1 & 0.354 & 0.202 & 0.808 & 1.845 & 3.209 & 49.6 & 114.81 \\
\hline\rowcolor{headerrow}\multicolumn{9}{l}{\textbf{35\%}} \\
OmniPrune & 248.6 & 0.353 & 1.794 & 0.646 & 1.940 & 4.733 & 53.8 & 114.83  \\
OmniPack & 257.9 & 0.353 & 0.389 & 1.681 & 1.995 & 4.418 & 54.9 & 120.34  \\
random (mean) & 246.4 & 0.354 & 0.098 & 0.635 & 3.105 & 4.191 & 85.3 & 114.82  \\
\rowcolor{oursrow}SOC (Ours) & 256.6 & 0.356 & 0.195 & 0.677 & 1.748 & 2.975 & 47.5 & 114.85 \\
\hline\rowcolor{headerrow}\multicolumn{9}{l}{\textbf{20\%}} \\
OmniPrune & 154.7 & 0.360 & 1.872 & 0.288 & 2.171 & 4.691 & 57.8 & 114.99  \\
OmniPack & 157.7 & 0.353 & 0.376 & 0.781 & 3.523 & 5.034 & 98.6 & 121.88  \\
random (mean) & 155.1 & 0.356 & 0.068 & 0.310 & 2.937 & 3.671 & 81.3 & 115.00  \\
\rowcolor{oursrow}SOC (Ours) & 158.2 & 0.357 & 0.120 & 0.355 & 1.888 & 2.721 & 51.1 & 115.00 \\
\hline\rowcolor{headerrow}\multicolumn{9}{l}{\textbf{15\%}} \\
OmniPrune & 130.9 & 0.355 & 1.881 & 0.232 & 2.196 & 4.664 & 61.2 & 114.99  \\
OmniPack & 133.8 & 0.356 & 0.374 & 0.560 & 2.312 & 3.602 & 63.1 & 122.16  \\
random (mean) & 131.6 & 0.354 & 0.051 & 0.228 & 2.376 & 3.010 & 66.5 & 115.00  \\
\rowcolor{oursrow}SOC (Ours) & 132.8 & 0.347 & 0.070 & 0.245 & 1.919 & 2.582 & 52.0 & 114.97 \\
\hline\end{tabularx}
\caption{\textbf{LongVALE and LongShOT inference costs across retention budgets.}}
\label{tab:appendix-longvale-efficiency-complete}
\label{tab:appendix-longshot-efficiency-complete}
\end{table}

\noindent\begin{minipage}[t]{0.49\textwidth}
\vspace{0pt}
\captionsetup{type=figure}
    \centering
    \includegraphics[width=0.72\linewidth]{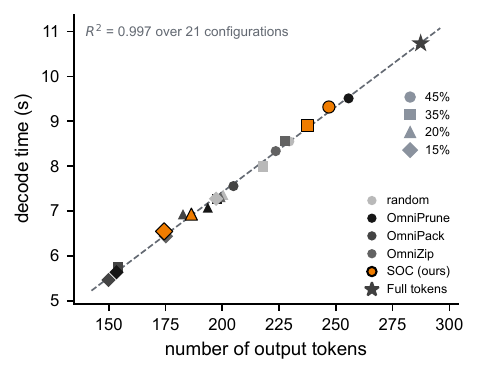}
    \caption{\textbf{LongVALE decode time follows output length}, so longer outputs take longer time even with shorter prefill latency.}
    \label{fig:longvale-decode}

\end{minipage}\hfill
\begin{minipage}[t]{0.49\textwidth}
\vspace{0pt}
\small
Figure~\ref{fig:longvale-decode} shows that decoding time closely follows output length across configurations.
At 35\% retention, SOC generates 237.6 tokens versus OmniPack's 154.1, resulting in longer decoding (8.898 versus 5.759\,s) despite shorter prefill (1.099 versus 1.218\,s).
Thus, total latency reflects both processing efficiency and response length; while prefill time provides an unaffected comparison.
\end{minipage}
\clearpage
\onecolumn

\section{More Visualization Results}
\label{sec:appendix-visualizations}

\begin{figure}[H]
\centering
\includegraphics[width=0.49\textwidth,height=0.78\textheight,keepaspectratio]{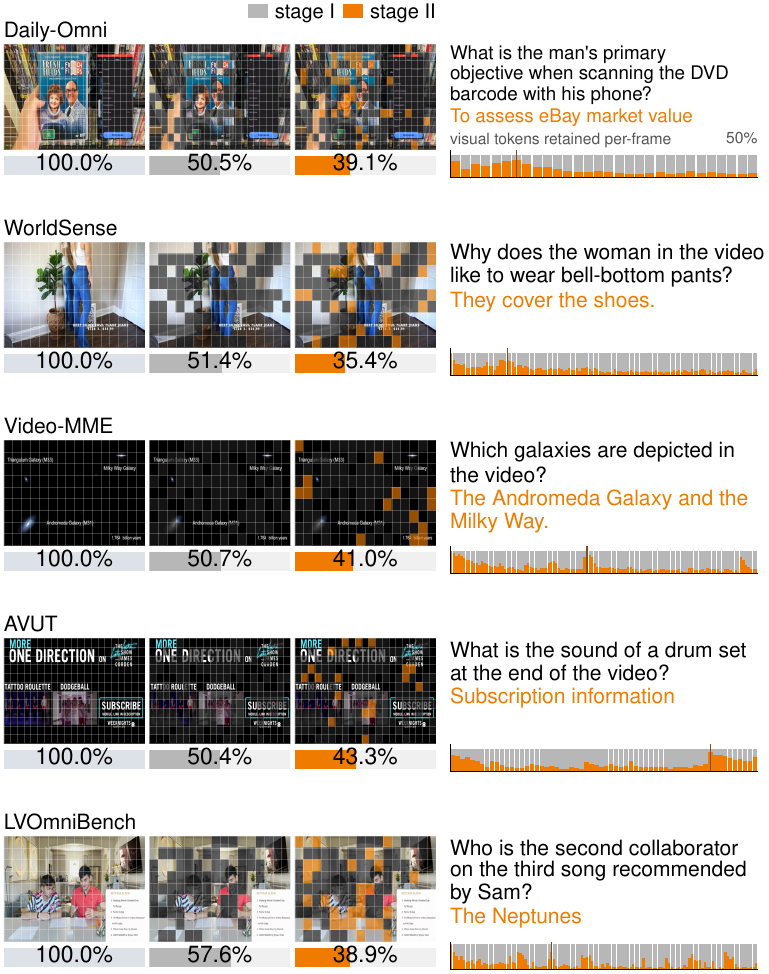}\hfill
\includegraphics[width=0.49\textwidth,height=0.78\textheight,keepaspectratio]{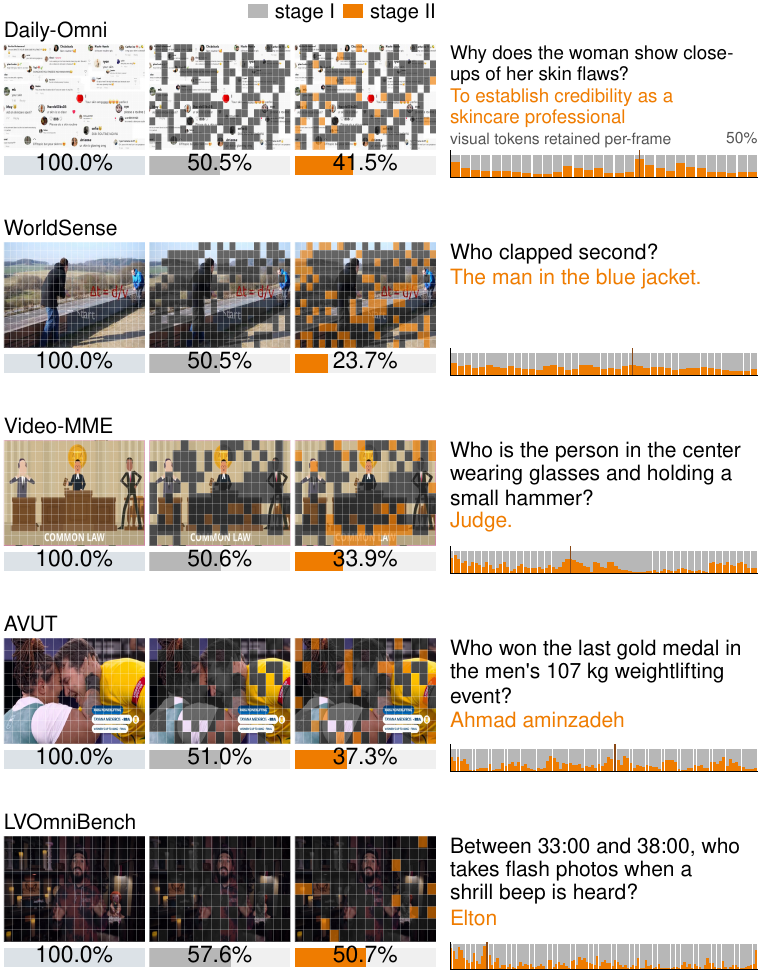}
\caption{\textbf{Additional two-stage token-selection examples.} 
Each row presents the question, input frame, and selections after stages I and II. 
Histograms show per-frame visual-token retention and the vertical marker identifies the displayed frame.}
\label{fig:appendix-token-vis}
\end{figure}

We provide ten additional examples from Daily-Omni, WorldSense, Video-MME, AVUT, and LVOmniBench in Figure~\ref{fig:appendix-token-vis}. 
Each example shows the input frame, token selection after both stages, and per-frame retention across the video.

\clearpage
\section{Attention Pattern Analyses}
\label{sec:appendix-attention-analysis}
\begin{figure}[H]
\centering
\includegraphics[width=\textwidth]{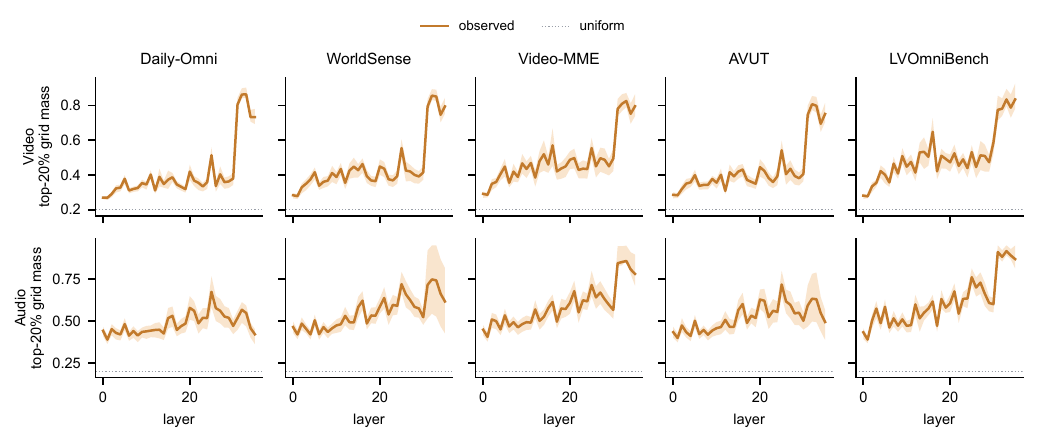}
\caption{\textbf{Temporal attention is broadly distributed in shallow layers and concentrates in deep layers for both video and audio.} 
Curves show the attention mass captured by the top-20\% temporal windows for video (top) and audio (bottom). 
The degree of concentration varies across benchmarks, while the overall pattern is consistent.}
\label{fig:attention-layers}
\end{figure}

\begin{figure}[H]
\centering
\includegraphics[width=\textwidth]{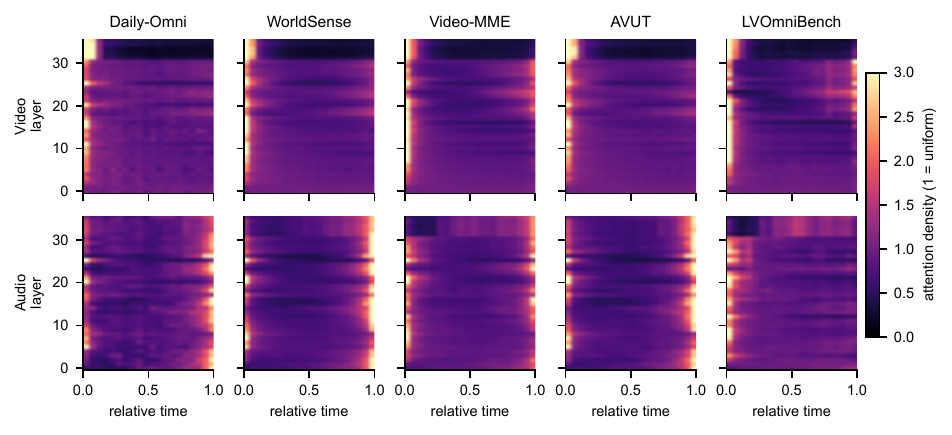}
\caption{\textbf{Video and audio exhibit temporal attention sinks at different positions of the sequence.} 
Heatmaps show mean attention density across layers and relative time on five benchmarks. 
Attention concentrates in deeper layers for both modalities. 
Video attention exhibits a sink primarily at the beginning of the sequence, while audio attention exhibits sink mostly at the end.}
\label{fig:attention-profiles}
\end{figure}

Figures~\ref{fig:attention-layers} and~\ref{fig:attention-profiles} extend the analysis in Section~\ref{sec:analysis-temporal} to all five QA benchmarks and to audio tokens.
For both modalities, temporal attention is more broadly distributed in shallow layers and concentrates in deeper layers, with a consistent trend across benchmarks despite variation in its strength.
At layer 0, the top 20\% of windows capture 27--29\% of video attention and 43--46\% of audio attention
, indicating a stronger early positional bias for audio.
The mean temporal profiles reveal attention sinks primarily near the start of the video and near the end of the audio.

\twocolumn

\section{Alternative Compression Designs}
\label{sec:appendix-alternatives}

\subsection{Cross-modal Token Pairing}
\label{sec:analysis-coupling}

\begin{table}[!ht]
\centering\scriptsize
\setlength{\tabcolsep}{3pt}\renewcommand{\arraystretch}{1.05}

\begin{tabular*}{\linewidth}{@{\extracolsep{\fill}}lrrrrrr@{}}
pairing budget & 0\% & 20\% & 40\% & 60\% & 80\% & 100\% \\
\hline
accuracy & \cellcolor{fullrow}62.0 & 61.9 & 61.0 & 60.7 & 57.6 & 56.5 \\
$\Delta$ & -- & $-0.1$ & $-1.0$ & $-1.3$ & $-4.3$ & $-5.5$ \\
overlap with audio saliency & -- & 6.6 & 12.5 & 18.4 & 24.0 & 31.0 \\
overlap with visual saliency & -- & 1.3 & 2.6 & 3.8 & 5.1 & 6.3 \\
\hline
\end{tabular*}

\caption{\textbf{Token pairing hurts performance and is not informative.} 
Cross-modal similarity tends to degrade accuracy and selects tokens with limited overlap with saliency-based choices.}
\label{tab:analysis-coupling}
\vspace{-8pt}
\end{table}

We retain 35\% of multimodal tokens and 45\% of audio tokens, varying retained audio tokens selected through pairing from 0\% to 100\%.
Audio-visual tokens are paired using cosine similarity of LLM hidden states within each frame.
Table~\ref{tab:analysis-coupling} shows that replacing saliency-selected tokens with cross-modal pairs progressively degrades accuracy.
Its limited overlap with saliency-based selection suggests that pairing may discard useful information.

\subsection{Token Fusion}
\label{sec:appendix-token-fusion}
We compare two visual token fusion methods with SOC on Daily-Omni using Qwen2.5-Omni-3B, retaining 50\%, 25\%, and 10\% visual tokens while keeping all audio tokens. 
FrameFusion~\cite{fu2025framefusion} averages tokens at the same spatial position across adjacent frames. 
AIM~\cite{zhong2025aim} merges similar tokens within each frame. 
Table~\ref{tab:token-fusion} shows that encoder saliency selection outperforms both token fusion methods.

\begin{table}[H]
\centering
\footnotesize
\begin{tabularx}{\columnwidth}{Xrrr}
 & 50\% & 25\% & 10\% \\
\hline
FrameFusion-style & 60.2 & 57.6 & 54.2 \\
AIM-style & 60.9 & 59.9 & 57.1 \\
\rowcolor{oursrow}SOC-style & \textbf{63.2} & \textbf{60.8} & \textbf{58.1} \\
\hline
\end{tabularx}
\caption{\textbf{Simple encoder saliency exceeds token fusion.}}
\label{tab:token-fusion}
\end{table}

\section{More Ablation Studies}
\label{sec:appendix-more-ablations}

\subsection{Contextual Refinement Location}
\label{sec:appendix-refinement-layer}
Our default setting applies stage II at the middle layer.
We vary the refinement layer $k$ on Qwen2.5-Omni-3B, using attention scores from layer $k-1$.
Refinement uses text queries and retains 45\% of audio tokens.
We fix $r_1:r_2=5:2$ and scale both ratios to maintain a layer-average audio-visual retention of $(kr_1+(L-k)r_2)/L=35\%$.
The three positions use $(r_1,r_2)=(63.64\%,25.45\%)$, $(50\%,20\%)$, and $(41.18\%,16.47\%)$, respectively.
Table~\ref{tab:appendix-refinement-layer} shows that middle-layer refinement performs best on four benchmarks and trails the best result on AVUT by only 0.12 points.

\begin{table}[H]
\centering
\footnotesize
\setlength{\tabcolsep}{2pt}
\begin{tabularx}{\columnwidth}{l*{5}{>{\raggedleft\arraybackslash}X}}
$k$ & Daily & World & Video & AVUT & LVOmni \\
\hline
\rowcolor{oursrow}18 (default) & \textbf{62.91} & \textbf{45.81} & \textbf{62.11} & 62.80 & \textbf{33.23} \\
9 & 62.32 & 45.08 & 60.52 & 61.59 & 31.66 \\
27 & 62.57 & 45.08 & 61.52 & \textbf{62.92} & 32.25 \\
\hline
\end{tabularx}
\caption{\textbf{Middle-layer refinement generally yields the best performance.}
Results on 3B model at 35\% total token retention.\label{tab:appendix-refinement-layer}}
\end{table}

\subsection{Refinement Budget Split}
\label{sec:appendix-schedules}
\begin{table}[H]
\centering
\scriptsize
\renewcommand{\arraystretch}{1.12}
\setlength{\tabcolsep}{3pt}
\resizebox{\columnwidth}{!}{%
\begin{tabular}{lcccccc}
retention & $r_1\!\to\!r_2$ & Daily & World & Video & AVUT & LVOmni \\
\hline
15\% & $17.5\!\to\!12.5$ & 57.73 & 43.76 & 58.11 & 59.63 & 33.33 \\
\rowcolor{oursrow}
 & $20\!\to\!10$ & \textbf{59.06} & \textbf{43.92} & 58.85 & 61.07 & \textbf{33.43} \\
 & $22.5\!\to\!7.5$ & 58.90 & 43.32 & 59.26 & \textbf{61.19} & 33.33 \\
 & $25\!\to\!5$ & 57.23 & 41.52 & \textbf{59.33} & 60.15 & 33.14 \\
 & $27.5\!\to\!2.5$ & 54.30 & 39.75 & 56.85 & 56.29 & 33.14 \\
\hline
20\% & $25\!\to\!15$ & 60.15 & 44.70 & 59.33 & 61.82 & 32.54 \\
 & $27.5\!\to\!12.5$ & 60.40 & \textbf{44.92} & 60.00 & 61.76 & 33.04 \\
\rowcolor{oursrow}
 & $30\!\to\!10$ & \textbf{60.90} & 44.70 & \textbf{60.56} & 62.05 & \textbf{33.83} \\
 & $32.5\!\to\!7.5$ & 60.23 & 44.42 & 59.67 & \textbf{62.17} & 33.63 \\
 & $35\!\to\!5$ & 58.23 & 42.62 & 59.37 & 60.90 & 32.64 \\
\hline
35\% & $40\!\to\!30$ & 62.82 & \textbf{45.81} & 61.26 & \textbf{62.98} & 32.64 \\
 & $45\!\to\!25$ & 62.82 & 45.74 & 62.07 & 62.40 & 32.25 \\
\rowcolor{oursrow}
 & $50\!\to\!20$ & \textbf{62.91} & \textbf{45.81} & \textbf{62.11} & 62.80 & 33.23 \\
 & $55\!\to\!15$ & 61.99 & 45.55 & 61.19 & 62.34 & \textbf{33.83} \\
 & $60\!\to\!10$ & 61.32 & 45.71 & 61.70 & 62.57 & 33.53 \\
\hline
\rowcolor{oursrow}
45\% & $50\!\to\!40$ & 63.41 & 45.46 & 61.85 & 63.09 & 32.84 \\
 & $55\!\to\!35$ & 63.58 & 45.81 & 62.30 & 62.92 & 32.84 \\
 & $60\!\to\!30$ & \textbf{63.83} & 46.31 & 62.30 & 62.98 & 33.14 \\
 & $65\!\to\!25$ & 62.99 & \textbf{46.60} & \textbf{62.63} & 63.09 & \textbf{33.33} \\
 & $70\!\to\!20$ & 62.57 & 46.56 & 62.04 & \textbf{63.32} & 33.14 \\
\hline
\end{tabular}%
}
\caption{Varying the budget of $r_1$ and $r_2$ across the four retention tiers on five QA benchmarks, Qwen2.5-Omni-3B.}
\label{tab:ablation-progressive-b}
\end{table}

We fix refinement at the middle layer ($k=18$) and vary $r_1$ and $r_2$ while keeping total token retention fixed.
Table~\ref{tab:ablation-progressive-b} reports results on Qwen2.5-Omni-3B.
The default splits perform well across benchmarks, although no single split is optimal in every setting.

\subsection{Comparison with More Baselines}
\label{sec:appendix-vision-only}
We evaluate VisionZip~\cite{yang2025visionzip} and DivPrune~\cite{alvar2025divprune} on Qwen2.5-Omni-3B at 35\% total audio-visual retention.
Table~\ref{tab:vision-only-3b} indicates SOC surpasses both baselines on all five benchmarks.

\begin{table}[h]
\centering
\footnotesize
\setlength{\tabcolsep}{2pt}
\begin{tabularx}{\columnwidth}{l*{5}{>{\centering\arraybackslash}X}}
 & Daily & World & Video & AVUT & LVOmni \\
\hline
VisionZip & 61.2 & 45.2 & 59.2 & 62.7 & 30.6 \\
DivPrune & 58.3 & 45.2 & 58.6 & 61.4 & 30.2 \\
\rowcolor{oursrow}SOC & \textbf{62.9} & \textbf{45.8} & \textbf{62.1} & \textbf{62.8} & \textbf{33.2} \\
\hline
\end{tabularx}
\caption{\textbf{SOC also exceeds encoder attention methods} (VisionZip~\cite{yang2025visionzip}) as well as \textbf{diversity selection} (DivPrune~\cite{alvar2025divprune}).}
\label{tab:vision-only-3b}
\end{table}

\end{document}